\documentclass[sigconf, nonacm]{acmart}

\renewcommand\footnotetextcopyrightpermission[1]{}

\AtBeginDocument{%
  }

\usepackage{enumitem}

\renewcommand\footnotetextcopyrightpermission[1]{} 
\usepackage{subcaption}
\usepackage{pifont}
\newcommand{\cmark}{\ding{51}}
\newcommand{\xmark}{\ding{55}}

\begin{document}

\title{Dream-Tac: A Unified Tactile World Action Model for Contact-Rich Robot Manipulation}

\author{Yunfan Lou}
\authornote{These authors contributed equally to this work.}
\affiliation{%
  \institution{Peking University}
  \city{Beijing}
  \country{China}
}

\author{Yifan Ye}
\authornotemark[1]
\authornote{Project leader.}
\affiliation{%
  \institution{Peking University}
  \city{Beijing}
  \country{China}
}

\author{Yankai Fu}
\authornotemark[1]
\affiliation{%
  \institution{Peking University}
  \city{Beijing}
  \country{China}
}

\author{Jun Cen}
\affiliation{%
  \institution{The Hong Kong University of Science and Technology}
  \city{Hong Kong}
  \country{China}
}

\author{Xiaowei Chi}
\affiliation{%
  \institution{The Hong Kong University of Science and Technology}
  \city{Hong Kong}
  \country{China}
}

\author{Yaoxu Lyu}
\affiliation{%
  \institution{Peking University}
  \city{Beijing}
  \country{China}
}

\author{Peidong Jia}
\affiliation{%
  \institution{Peking University}
  \city{Beijing}
  \country{China}
}

\author{Sirui Han}
\affiliation{%
  \institution{The Hong Kong University of Science and Technology}
  \city{Hong Kong}
  \country{China}
}

\author{Zhihe Lu}
\affiliation{%
  \institution{Nanjing University}
  \city{Nanjing}
  \country{China}
}

\author{Shanghang Zhang}
\authornote{Corresponding author.}
\affiliation{%
  \institution{State Key Laboratory of Multimedia Information Processing, School of Computer Science, Peking University}
  \city{Beijing}
  \country{China}
}
\email{shanghang@pku.edu.cn}

\renewcommand{\shortauthors}{Lou, Ye, Fu, et al.}


\begin{abstract}

World action models inherit the predictive capability of world models, enabling action generation to be guided by anticipated future observations. However, they rely primarily on vision and often fail in contact-rich manipulation, where critical cues arise from physical interaction. In this paper, we propose Dream-Tac, a unified Tactile-World Action Model that jointly models actions, future visual observations, and tactile dynamics. Specifically, Dream-Tac introduces (i) contact-gated visuotactile fusion to selectively integrate tactile signals and (ii) a contact-aware attention bias to better regulate cross-modal interactions during manipulation. To support real-time deployment, we further design a dual-level acceleration strategy, reformulating the contact-aware bias to preserve the fused attention path during training and introducing cache-based diffusion acceleration at inference, achieving up to 2.9$\times$ faster training and 1.8$\times$ faster inference. Across six contact-rich manipulation tasks, Dream-Tac improves action accuracy by 31.7\% on average, demonstrating the effectiveness of unified visuotactile world modeling.Code is available at \url{https://github.com/LYFCLOUDFAN/Dream-Tac}.
\end{abstract}

\begin{teaserfigure}
  \includegraphics[width=\textwidth]{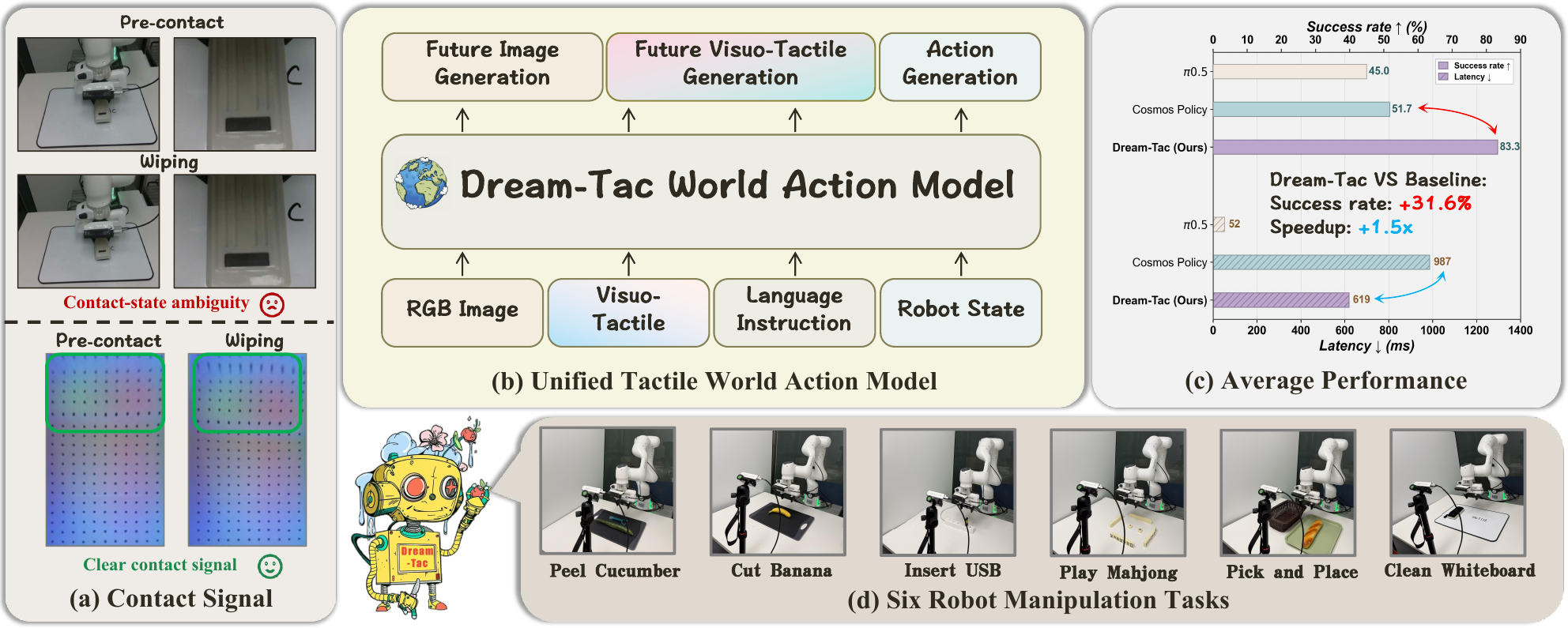}
  \caption{Overview of our work. (a) Comparison of RGB and tactile sensing for perceiving contact-state changes before and after contact. (b) Illustration of Dream-Tac, a unified tactile world action model, including its inputs and outputs. (c) Comparison with baseline methods on six contact-rich manipulation tasks. (d) Visualization of the settings for the six tasks.
  }
  \Description{Enjoying the baseball game from the third-base
  seats. Ichiro Suzuki preparing to bat.}
  \label{fig:teaser}
\end{teaserfigure}


\maketitle

\section{Introduction}

World models learn predictive representations of environmental dynamics by forecasting future observations\cite{cosmos2,cosmosworld,liao2025genie,cosmos-reason,V-JEPA}, thereby endowing agents with the ability to anticipate how the environment will evolve over time. Building on this paradigm, world action models further transfer such predictive knowledge into the decision-making process\cite{videovla,cosmos-policy,dreamzero}, allowing policies to inherit dynamics priors acquired from large-scale web-data pretraining and to use these priors to guide action generation. 

Although world action models have shown promising performance in general manipulation tasks~\cite{dreamzero, cosmos-policy, li2026causal}, they remain limited in contact-rich and fine-grained settings. This limitation stems largely from the fact that vision alone often fails to capture the physical interaction cues required for precise control, including contact states, local geometry, and fine-grained object properties~\cite{vtla,vla-touch,tactile-vla,omnivtla, fu2025cordvip}. As a result, visually conditioned policies continue to struggle in manipulation settings where success depends on accurately perceiving contact. Tactile sensing naturally addresses this gap by directly providing contact and local interaction signals that are ambiguous or entirely unavailable in RGB observations~\cite{tactile-review}. As shown in Fig.~\ref{fig:teaser} (a), RGB images do not provide a clear indication of contact, whereas tactile variations reveal subtle physical interaction cues that compensate for the lack of fine-grained contact perception in vision. This makes tactile sensing particularly important for manipulation tasks where accurate awareness of physical contact is critical~\cite{li2014localization}.

Incorporating tactile feedback into policy learning provides an effective pathway toward precise decision-making \cite{omnivtla, vtla, xue2025reactive, li2025adaptive}, particularly within world action models. Tactile signals are inherently temporal and event-driven, while world models are designed to capture temporal dynamics, making them naturally well-suited for modeling such interaction patterns \cite{higuera2026visuo}. However, these signals are sparse and transient in practice; 
long periods of stasis are often punctuated by brief, critical events like contact onset, slip, or release. 
Consequently, treating tactile tokens symmetrically with other modalities risks diluting the precise interaction cues most essential for successful manipulation. This raises a key question: how can tactile signals be incorporated into world action models in a selective and interaction-aware manner?


Motivated by these observations, we propose Dream-Tac, a unified tactile world action model that integrates tactile perception into a generative framework by jointly predicting future visual observations, robot actions, and future tactile observations. 
Built upon a pretrained video generative backbone \cite{cosmos2}, Dream-Tac extends world action modeling to contact-rich manipulation through a novel contact-aware attention bias. 
This mechanism mitigates the sparsity of tactile signals by adaptively amplifying the influence of touch only when contact dynamics become salient. 
Rather than treating tactile tokens uniformly, Dream-Tac prioritizes interaction-relevant signals, enabling tighter coupling between future visual states, tactile dynamics, and precise action generation. Extensive real-world experiments validate the effectiveness of out method, achieving the highest success rate among four strong baselines and outperforming Cosmos Policy \cite{cosmos-policy} by 31.6\%.

While these tactile-aware mechanisms enhance perception, they significantly increase the computational burden of the world action model. 
This challenge is particularly pronounced in world action models built upon Diffusion Transformers \cite{peebles2023scalable, peebles2023scalable}, which involve quadratic self-attention and iterative multi-step denoising. The inclusion of tactile sequences further increases the temporal and modality complexity, making real-time deployment more challenging.
To mitigate this, Dream-Tac incorporates a dual-level acceleration system. 
First, we optimize training by re-implementing the gated-bias attention with a FlashAttention-based formulation \cite{dao2022flashattention, dao2023flashattention2,wu2025flashbias}, achieving up to 2.94$\times$ training speedup. 
Second, we accelerate inference via diffusion-step caching, yielding a 1.8$\times$ speedup at test time.
These system-level optimizations are critical for ensuring Dream-Tac remains feasible for the high-frequency execution required in precise robotic manipulation.


In summary, our contributions are four-fold: 
\begin{itemize}[nosep]
\item We propose Dream-Tac, a unified generative world action model that jointly predicts future visual observations, tactile signals, and robot actions.
\item We propose a contact-aware attention mechanism that adaptively emphasizes tactile signals during salient interaction events.
\item We develop a dual-level acceleration system featuring a FlashAttention-based gated bias for efficient training and a diffusion-step caching strategy for faster inference.
\item Experiments on contact-rich manipulation tasks demonstrate that Dream-Tac significantly outperforms state-of-the-art baselines in success rate while maintaining competitive generation quality and execution efficiency.
\end{itemize}

\section{Related Work}
\subsection{Tactile in Robot Manipulation}
A wide range of works have explored incorporating tactile signals into robotic policies, demonstrating their importance for capturing contact and interaction cues in tasks such as grasping~\cite{calandra2018more, polic2019convolutional}, insertion~\cite{dong2021tactile, ma2019dense},  in-hand manipulation~\cite{she2021cable, qi2023general} and tool use~\cite{suh2022seed, aoyama2025few}.

More recently, visuo-tactile perception has been incorporated into end-to-end manipulation policies~\cite{zhang2025ta, vla-touch, vtla, heng2025vitacformer}. For example, RDP~\cite{xue2025reactive} combines a slow diffusion policy with fast tactile feedback for reactive control. AdapTac~\cite{li2025adaptive} adaptively fuses 3D information and tactile features with force-guided attention for dexterous manipulation. Tactile-VLA~\cite{tactile-vla} incorporates tactile sensing into a unified VLA framework for contact-rich manipulation. Another line of work focuses on tactile representation learning and visuo-tactile pretraining, aiming to learn general contact-aware features that transfer across downstream manipulation tasks~\cite{wu2025canonical, zhu2025touch, higuera2025tactile}. While these studies have improved tactile-aware control and representation learning, the use of tactile signals for explicitly modeling future state evolution and contact dynamics remains underexplored.

In this work, we integrate tactile sensing into the world model and explicitly model the future evolution of both visual and tactile observations conditioned on robot actions. This allows the policy to capture contact dynamics more directly and supports more predictive contact-aware manipulation.
 
\subsection{World Action Model}
World models learn predictive representations by forecasting future observations~\cite{du2023learning, hafner2019dream, hafner2023mastering, lou2026maskworldmodel}. Their recent success is partly driven by large-scale pretraining on web data, which enables them to capture broad knowledge of visual dynamics and scene evolution~\cite{wu2023unleashing, V-JEPA, chi2025wow}. These properties make world models particularly appealing for embodied AI, where understanding how the environment may evolve is closely tied to effective decision-making.

World Action Models (WAMs) build on this idea by bringing world modeling into policy learning~\cite{cosmos-policy, li2026causal, cen2025worldvla, cen2025rynnvla, zhang2025dreamvla, ye2025self}. Instead of predicting actions solely from the current context, WAMs leverage future visual prediction as an auxiliary structure or intermediate representation for action generation. Existing approaches generally fall into two categories. Imagine-then-execute methods first synthesize future visual trajectories and then use them for downstream control~\cite{li2026causal, zhou2025act2goal, hu2024video}. Joint modeling methods instead learn actions and future observations within a unified generative process~\cite{cosmos-policy, zhu2025unified, dreamzero}. Both directions share the same high-level goal: to exploit pretrained dynamics knowledge from world models to improve action understanding and generation.

Our method is most closely related to joint modeling WAMs, but differs in modality and focus. Rather than restricting world action modeling to vision alone, we incorporate tactile sensing into the predictive framework. Concretely, we formulate a unified visuotactile world action model that jointly predicts future visual observations, future tactile signals, and robot actions. This extension allows the model to capture contact dynamics beyond what is observable from vision, leading to stronger embodied representations for fine-grained manipulation.

\section{Method}
\subsection{Problem Formulation}
We study real-time robot manipulation conditioned on visual observations, tactile observations, and language instructions. Let \(o\) denote the current visual observation, \(x\) denote the current tactile observation, and \(l\) denote the task instruction. The goal is to predict an action chunk \(a_{1:H}\) over a horizon of \(H\) steps.

A standard visuomotor policy directly models actions from the current observation and language instruction:
\begin{equation}
p(a_{1:H} \mid o, l).
\label{eq:policy}
\end{equation}

In parallel, a world model focuses on predicting future observations. Let \(v_{1:T}\) denote future visual observations over a horizon of \(T\) steps. A standard world model captures the conditional distribution
\begin{equation}
p(v_{1:T} \mid o, l),
\label{eq:wm}
\end{equation}
which models how the visual scene evolves given the current context.

Building on these two formulations, a world action model combines action prediction and future observation prediction into a unified framework. Specifically, it jointly models
\begin{equation}
p(a_{1:H}, v_{1:T} \mid o, l),
\label{eq:wam_joint}
\end{equation}
or equivalently factorizes the joint distribution as
\begin{equation}
p(a_{1:H}, v_{1:T} \mid o, l)
=
p(v_{1:T} \mid o, l)\, p(a_{1:H} \mid o, l, v_{1:T}),
\label{eq:wam_factorized}
\end{equation}
where future visual prediction provides predictive structure for action generation.

However, in contact-rich manipulation, vision alone is often insufficient to capture physical interaction cues. To address this limitation, we introduce Dream-Tac, an enhanced world action model that incorporates tactile sensing into both the conditioning context and the prediction targets. Let \(x_{1:T}\) denote future tactile observations. Dream-Tac jointly models
\begin{equation}
p(a_{1:H}, v_{1:T}, x_{1:T} \mid o, x, l),
\label{eq:tacwam_joint}
\end{equation}
thereby extending standard world action modeling from visual future prediction to joint visuo-tactile future prediction. This formulation enables the model to reason jointly about action generation, future scene evolution, and future contact dynamics in a unified framework.

\subsection{Dream-Tac Architecture}

\begin{figure*}[t]
    \centering
    \includegraphics[width=\textwidth]{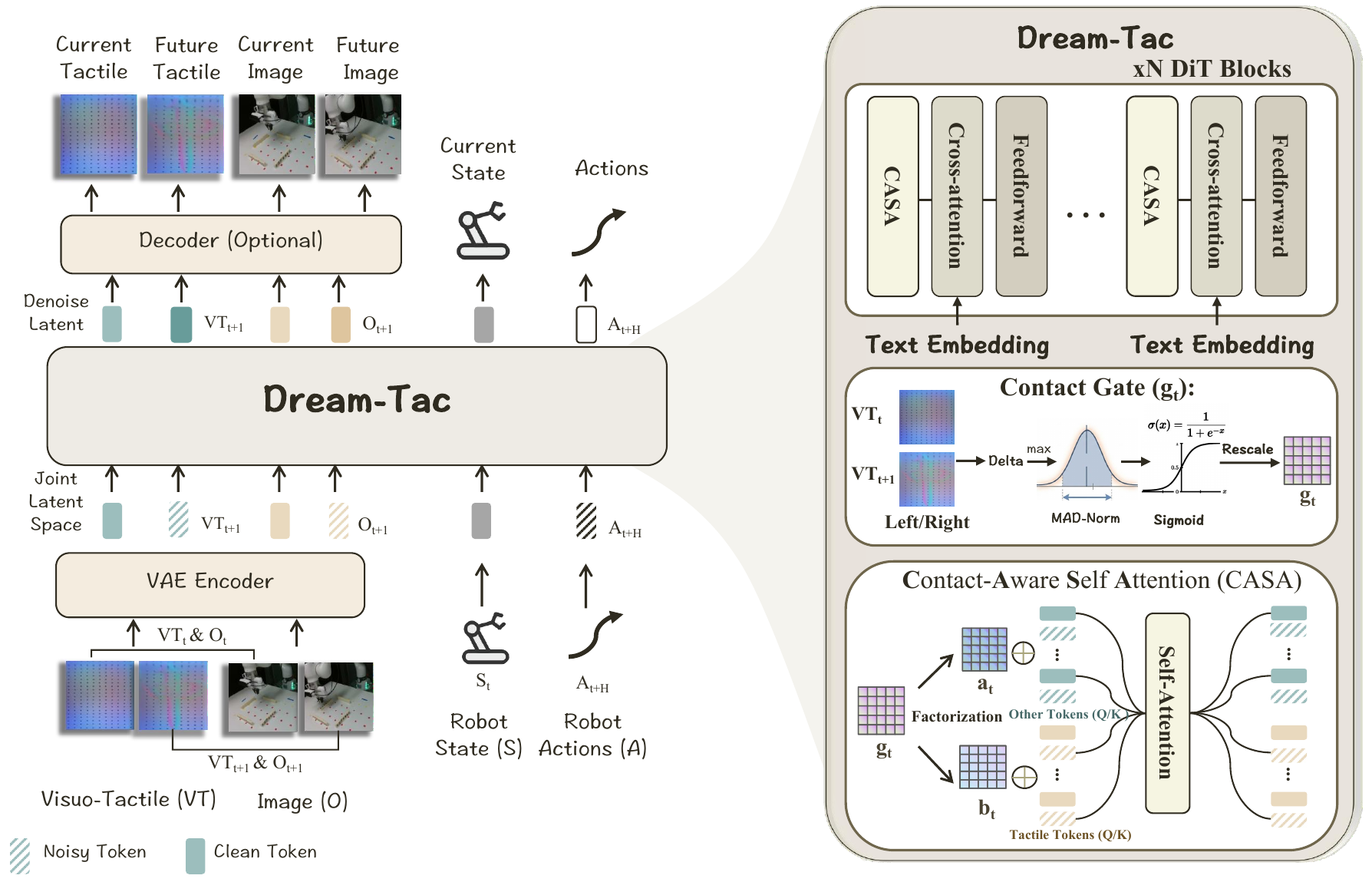}
    \caption{\textbf{Overview of Dream-Tac.} Dream-Tac is a unified tactile world action model that jointly predicts future tactile observations, future visual observations, and robot actions in a shared latent space. Left: multimodal inputs, including visuo-tactile observations, visual observations, robot states, and action tokens, are encoded into a joint latent representation and processed by a shared backbone, optionally followed by decoding for future prediction. Right: the Dream-Tac backbone is built on DiT blocks and incorporates the proposed contact-aware self-attention (CASA). A contact gate $g_t$ is computed from frame-to-frame tactile changes and used to rescale the tactile attention bias, enabling the model to emphasize tactile information during salient interaction events.}
    \label{fig:main_arch}
\end{figure*}

As illustrated in Fig.~\ref{fig:main_arch},Dream-Tac is built on top of a pretrained video Diffusion Transformer (DiT) \cite{cosmos2,DIT}, which serves as the backbone for world modeling. We reuse its web-data-pretrained T5 text encoder and video VAE. Task language is encoded by the T5 encoder and provided to all tokens through cross-attention, while visual observations are encoded into latent video tokens by the pretrained VAE.

On top of this backbone, we incorporate robot states, action chunks, and tactile observations into the same latent modeling framework. Following Cosmos-Policy\cite{cosmos-policy}, robot states and actions are represented as padded latent-frame tokens and inserted into the video backbone for joint denoising. To capture contact-aware interaction, tactile observations are also mapped into the same latent space through the same pretrained VAE. This unified latent representation allows Dream-Tac to jointly model visual dynamics, tactile dynamics, and action generation within a single diffusion transformer, while preserving the stability and priors of the pretrained backbone.

An important advantage of this design is that tactile signals become directly accessible to the shared transformer. Through bidirectional self-attention, action tokens can condition not only on future visual observations but also on fine-grained changes in future tactile signals. This enables Dream-Tac to leverage anticipated contact dynamics for precise manipulation, especially in contact-rich tasks where vision alone is insufficient. Moreover, as validated in our experiments, the pretrained VAE already provides a meaningful latent separation of tactile patterns associated with different actions, without requiring additional tactile-specific pretraining.

\subsection{Contact-Aware Self Attention}

Because Dream-Tac jointly models robot state, visual observations, and tactile observations in a shared backbone, it attends to all three modalities uniformly at each timestep. However, tactile signals in contact-rich manipulation are often sparse (in general) and transient: they remain nearly constant for long periods, while contact onset, slip, or release causes sharp local variations. Under symmetric attention, non-contact timesteps may still attend to tactile tokens, causing weakly relevant tactile information to accumulate and weakening the model’s sensitivity to contact-state changes. To address this, we introduce a gated attention mechanism that amplifies tactile signals during contact events.

\textbf{Attention with a gated, structured logit bias.}
We retain the standard linear projections and value aggregation, and add a scalar-gated additive bias to the attention logits so that non-tactile queries can increase attention to tactile keys according to a data-derived contact-event strength.
Let $i,j$ index flattened tokens, and let $M_i\in\{0,1\}$ indicate whether token $i$ belongs to the tactile token.
For each input sequence and timestep $t$, we compute a gate $g_t\in[g_{\min},g_{\max}]$ (defined below). The attention logits are
\begin{equation}
\label{eq:casa_logit}
\mathrm{logit}_{ij}
=
\frac{\mathbf{q}_i^\top \mathbf{k}_j}{\sqrt{d}}
\;+\;
\alpha\, g_t\,(1-M_i)\,M_j,
\end{equation}
where $d$ is the head dimension and $\alpha>0$ controls the bias magnitude.
The bias is active only when a non-tactile query ($M_i=0$) attends to a tactile key ($M_j=1$), yielding an asymmetric inductive bias that encourages action, proprioceptive, and visual tokens to access tactile information when contact dynamics are salient, while leaving tactile-to-tactile interactions unchanged.
Attention weights are computed as $a_{ij}=\mathrm{softmax}_j(\mathrm{logit}_{ij})$, followed by standard value aggregation.

\textbf{Gate computation.}
The gate $g_t$ measures the magnitude of tactile change between consecutive timesteps along a trajectory, using raw tactile RGB frames without introducing an additional learned gating network.
Let $I^{L}_{t}, I^{R}_{t}\in\{0,\ldots,255\}^{H\times W\times 3}$ denote the left and right tactile images obtained by the sensors mounted on the gripper at time $t$.
For each view, we compute the normalized mean absolute difference to the previous frame:
\begin{equation}
\label{eq:casa_delta}
\begin{aligned}
\delta^{L}_{t}
&=
\frac{1}{255}\,\mathbb{E}_{p,c}\!\left[\left|I^{L}_{t}(p,c)-I^{L}_{t-1}(p,c)\right|\right],\\
\delta^{R}_{t}
&=
\frac{1}{255}\,\mathbb{E}_{p,c}\!\left[\left|I^{R}_{t}(p,c)-I^{R}_{t-1}(p,c)\right|\right].
\end{aligned}
\end{equation}
where $(p,c)$ indexes spatial locations and RGB channels.
We define the per-timestep event strength as
\begin{equation}
\label{eq:casa_rho}
\rho_t = \max\!\left(\delta^{L}_{t},\delta^{R}_{t}\right),
\end{equation}
so that a salient change in either tactile view increases the gate.
For the initial timestep, where no predecessor frame is available, we set $\rho_0=0$.

To suppress benign sensor noise while preserving salient contact events, we map $\rho_t$ to a bounded gate using a fixed robust normalization:
\begin{equation}
\label{eq:casa_gate}
\begin{aligned}
z_t 
&= k\,\frac{\rho_t - m}{s+\epsilon},\\
\tilde g_t 
&= \sigma(z_t)=\frac{1}{1+e^{-z_t}},\\
g_t 
&= g_{\min} + (g_{\max}-g_{\min})\,\tilde g_t,
\end{aligned}
\end{equation}
where $m$ and $s$ are fixed reference location and scale hyperparameters (used in a median-/MAD-style normalization, rather than estimated from each dataset), $k$ controls the sigmoid sharpness, and $\epsilon$ ensures numerical stability. In our implementation, $(m,s,k,\epsilon)=(0.002,0.001,4,10^{-6})$, and $z_t$ is clipped to $[-30,30]$ before the sigmoid to avoid extreme saturation.
As a result, small background fluctuations yield $g_t\approx g_{\min}$, whereas pronounced frame-to-frame tactile changes push $g_t$ toward $g_{\max}$, strengthening the directed bias in Eq.~\eqref{eq:casa_logit} when contact dynamics are more likely to be changing.
Because the gate is computed independently for each sample and timestep from observed tactile inputs alone, it can be used during both training and inference without introducing privileged future information.

\subsection{Training Objective}

Dream-Tac is trained with the same latent denoising objective as the underlying pretrained video model. 
Given the current visual observation \(o\), current tactile observation \(x\), and language instruction \(l\), 
we jointly model future visual observations \(v_{1:T}\), future tactile observations \(x_{1:T}\), and action chunk \(a_{1:H}\) 
within a unified latent sequence. Specifically, visual observations and tactile observations are encoded into latent tokens 
using the pretrained VAE, while robot actions are represented as latent-frame tokens following latent injection. Let
$y = \{z^{v}_{1:T}, z^{x}_{1:T}, z^{a}_{1:H}\}$
denote the target latent variables, where \(z^{v}_{1:T}\) are the latent tokens of future visual observations, 
\(z^{x}_{1:T}\) are the latent tokens of future tactile observations, and \(z^{a}_{1:H}\) are the latent action tokens. 
During training, we sample Gaussian noise \(\epsilon \sim \mathcal{N}(0, I)\) and a noise level \(\sigma \sim p(\sigma)\), 
and construct the corrupted latent sequence
\begin{equation}
\tilde{y} = y + \sigma \epsilon.
\label{eq:noisy_latent}
\end{equation}

The diffusion transformer is trained to predict the denoising target conditioned on the clean prefix context 
(consisting of current observations and language) together with the corrupted latent sequence. 
Using the standard denoising objective of the pretrained backbone, the training loss is written as
\begin{equation}
\mathcal{L}_{\mathrm{denoise}}
=
\mathbb{E}_{y,\epsilon,\sigma}
\Big[
\| f_{\theta}(\tilde{y}, \sigma, o, x, l) - \epsilon \|_2^2
\Big],
\label{eq:denoise_loss}
\end{equation}
where \(f_{\theta}\) denotes the Dream-Tac backbone.

Since Dream-Tac jointly models actions, future images, and future tactile signals in the same latent space, 
this denoising objective simultaneously supervises action generation, future visual prediction, and future tactile prediction. 
For clarity, the objective can also be decomposed into modality-specific terms:
\begin{equation}
\mathcal{L}
=
\mathcal{L}_{\mathrm{act}}
+
\lambda_v \mathcal{L}_{\mathrm{img}}
+
\lambda_t \mathcal{L}_{\mathrm{tac}},
\label{eq:total_loss}
\end{equation}
where \(\mathcal{L}_{\mathrm{act}}\), \(\mathcal{L}_{\mathrm{img}}\), and \(\mathcal{L}_{\mathrm{tac}}\) correspond to the denoising losses 
on action, future visual, and future tactile latent tokens, respectively, and \(\lambda_v\) and \(\lambda_t\) balance the three objectives.

\subsection{Efficiency Design}
To make Dream-Tac practical for real-time robot deployment \cite{ye2025token}, we introduce a system-level acceleration module. Since vanilla FlashAttention \cite{dao2022flashattention,dao2023flashattention2} does not natively support our gated-bias attention, we follow FlashBias \cite{wu2025flashbias} to re-implement the proposed contact-aware gated bias attention with a low-rank bias formulation on top of FlashAttention. This optimization significantly improves efficiency, reducing training latency on NVIDIA H200 GPUs from 97 ms to 29 ms per iteration.
At inference time, the video backbone must process the full model for each denoising step, making multi-step diffusion particularly expensive for real-time control. For example, with 10 denoising steps, our method runs at only 5Hz on an A800 GPU. To address this bottleneck, we introduce a diffusion-step cache strategy\cite{dreamzero} that reduces the number of effective inference steps to 2 while preserving performance. 

\section{Experiment}
We evaluate Dream-Tac on real-world contact-rich manipulation in terms of effectiveness, efficiency, generalization, and ablation. Specifically, we compare against strong baselines on six real-world contact-rich manipulation tasks, measure success rates, training and inference efficiency, test robustness under out-of-distribution environment variations, and perform ablations to isolate the roles of tactile fusion, contact-aware attention bias, and the proposed acceleration strategy.

\subsection{Experimental Setups}
We evaluate Dream-Tac on a Franka Emika Panda robot in a real-world tabletop setting.
The perception system consists of two synchronized Intel RealSense D435i RGB cameras, including a fixed third-person camera and a wrist-mounted camera, together with two Xense Photon tactile sensors mounted on the gripper fingertips.
This setup provides complementary global visual context, egocentric close-range observations, and contact-rich tactile feedback.

\noindent\textbf{Task Description:} We consider six language-conditioned manipulation tasks.
\begin{enumerate}
    \item \textbf{Pick Baguette}: the robot moves a baguette from a plate into a basket; success requires the baguette to be fully inside the basket without noticeably moving the basket.
    \item \textbf{Insert USB}: the robot inserts a USB device into a power strip; success requires full insertion.
    \item \textbf{Clean Whiteboard}: the robot uses an eraser to remove handwritten marks from the whiteboard and then return the eraser to its original position; success requires the writing to become invisible to the human eye and the eraser to be placed back correctly.
    \item \textbf{Peel Cucumber}: the robot uses a peeler to peel a cucumber; success requires producing a peeled strip longer than 5\,cm.
    \item \textbf{Play Mahjong}: the robot determines whether the observed tile results in a winning hand; if yes, it pushes the tiles down, otherwise it places the tile at the center of the table; success requires both correct judgment and correct execution.
    \item \textbf{Cut Banana}: the robot cuts a banana with a knife; success is defined as a cut depth exceeding 5\,cm.
\end{enumerate}

\noindent\textbf{Baselines.} We compare Dream-Tac with four strong baselines: (1) $\pi_0$ \cite{black2024pi_0}: a VLA flow model for general robot
control; (2) $\pi_{0.5}$ \cite{intelligence2504pi0}, an enhanced VLA flow model for general physical reasoning and long-horizon manipulation; (3) ForceVLA \cite{yu2025forcevla}: a multimodal VLA model integrating force-torque sensing for contact-rich manipulation; (4) Cosmos Policy \cite{cosmos-policy}: a world action model leveraging physical commonsense for complex task execution.

\noindent\textbf{Evaluation protocol.}
For each method on each task, we conduct 20 real-world evaluation trials, and all baselines are evaluated under the same trial budget.
For each method, we report the success rate of the best-performing checkpoint, selected according to the same rule across methods.
All methods are evaluated under the same task definitions, success criteria, and real-world setup.

\begin{figure*}[t]
    \centering
    \includegraphics[width=\textwidth]{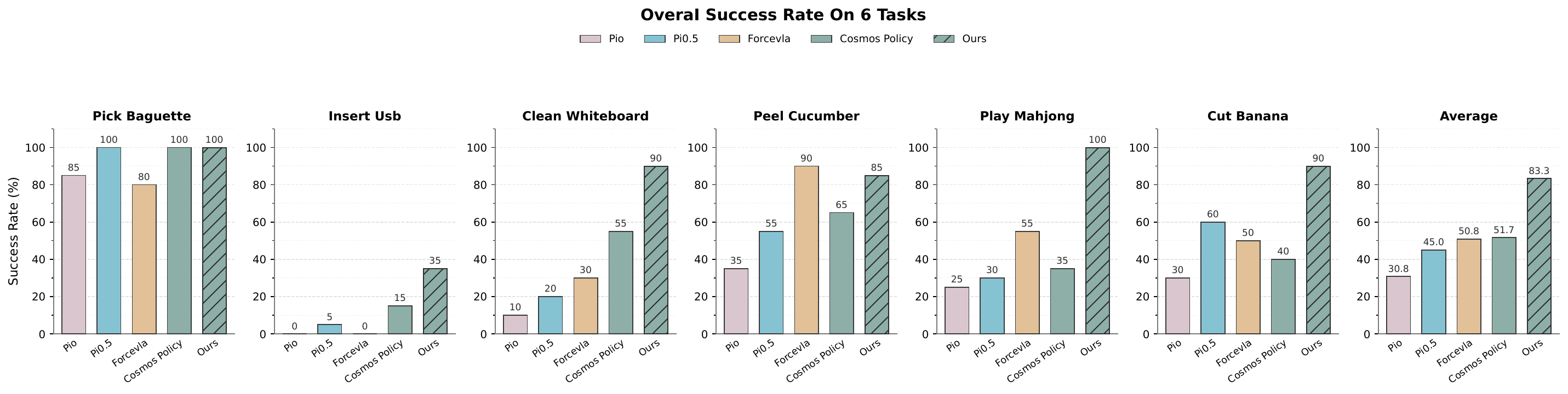}
    \caption{For each method on each task, we conduct 20 real-world evaluation trials, and all baselines are evaluated under the same trial budget.For each method, we report the success rate of the best-performing checkpoint, selected according to the same rule across methods.
All methods are evaluated under the same task definitions, success criteria, and real-world setup.}
    \label{fig:overall_sr_6tasks}
\end{figure*}

\subsection{Evaluations}
\subsubsection{Performance on Real-World Experiments}

We conduct experiments on six real-world manipulation tasks.
As shown in Fig.~\ref{fig:overall_sr_6tasks} and Table.~\ref{tab:ablation_sr}, Dream-Tac achieves the best overall performance, with an average success rate of \textbf{83.3\%}, substantially outperforming Cosmos-Policy (51.7\%), ForceVLA (50.8\%), $\pi_{0.5}$ (45.0\%), and $\pi_0$ (30.8\%).

Dream-Tac attains the highest success rate on five of the six tasks, including \textbf{Pick Baguette} (100\%), \textbf{Insert USB} (35\%), \textbf{Clean Whiteboard} (90\%), \textbf{Play Mahjong} (100\%), and \textbf{Cut Banana} (90\%).

On \textbf{Peel Cucumber}, Dream-Tac achieves \textbf{85\%}, remaining competitive while slightly below ForceVLA.

Notably, for relatively coarse manipulation tasks such as \textbf{Pick Baguette}, most baselines already achieve fairly high success rates, suggesting that these tasks can often be solved with strong visual perception and basic motion control alone.
In contrast, Dream-Tac shows its largest advantage on tasks requiring precise contact-rich interaction and fine-grained adjustment, such as \textbf{Insert USB} and \textbf{Cut Banana}, where it achieves state-of-the-art performance by a clear margin.
These tasks demand accurate reasoning about subtle contact transitions, spatial alignment, and forceful interaction during execution, making them particularly challenging for methods that rely only on current observations or reactive policies.

The \textbf{Play Mahjong} task further highlights the benefit of tactile reasoning under severely limited visual input.
During Mahjong play, we intentionally construct a fully occluded setting in which visual observations are completely blocked, forcing the model to rely almost entirely on tactile feedback to determine the next action pattern.
In this setting, we compare ForceVLA with our two tactile-dependent variants.
We find that our method remains highly robust, which we attribute in part to the proposed attention bias mechanism.
By encouraging the policy to attend more strongly to tactile cues, the attention bias alleviates overfitting to image observations and improves the model's ability to act under visual deprivation.
In contrast, ForceVLA still exhibits a tendency to rely on visual observations and robot state priors in some cases, which limits its robustness when the visual modality becomes uninformative or unavailable.
These results suggest that Dream-Tac learns a more genuinely tactile-grounded policy, rather than treating tactile input as only an auxiliary signal.


\subsubsection{Ablation Studies}

We conduct ablation studies to isolate the contributions of tactile fusion and contact-aware attention bias.
Specifically, we consider three variants:
\textbf{(1) Visual WAM}, which uses visual observations only;
\textbf{(2) Visuo-tactile WAM}, which incorporates tactile observations but uses standard self-attention without the proposed bias;
and \textbf{(3) Visuo-tactile WAM + Bias}, which further introduces the proposed contact-aware attention bias on top of tactile fusion.
These three settings are consistent with the training-efficiency comparison shown in the left panel of Fig.~\ref{fig:ab_step}.

Table~\ref{tab:ablation_sr} reports the average success rate over the six real-world tasks.
Compared with Visual WAM, incorporating tactile observations in Visuo-tactile WAM improves the average success rate from 51.7\% to 74.2\%, showing that tactile fusion is the main source of performance gain.
Further adding the proposed attention bias increases the average success rate to 83.3\%, indicating that contact-aware attention bias provides additional benefits by better exploiting tactile-relevant interaction cues.

\begin{table}[t]
\caption{Ablation on tactile fusion and contact-aware attention bias. Results are reported as average success rate (\%) over six real-world tasks.}
\label{tab:ablation_sr}
\centering
\small
\setlength{\tabcolsep}{5pt}
\begin{tabular}{lccc}
\toprule
Method & Tactile & Attn. Bias & Avg. SR (\%) \\
\midrule
Visual WAM                & \xmark & \xmark & 51.7 \\
Visuo-tactile WAM         & \cmark & \xmark & 74.2 \\
Visuo-tactile WAM + Bias  & \cmark & \cmark & \textbf{83.3} \\
\bottomrule
\end{tabular}
\end{table}

\subsubsection{Generalization Under Environment Variations}

We further evaluate Dream-Tac under four types of out-of-distribution environment variations: table height, spatial arrangement, object appearance, and background.
These variations are instantiated on four representative tasks: \textit{Peel Cucumber}, \textit{Pick Baguette}, \textit{Play Mahjong}, and \textit{Cut Banana}, respectively.

\begin{figure*}[t]
    \centering
    \includegraphics[width=\textwidth]{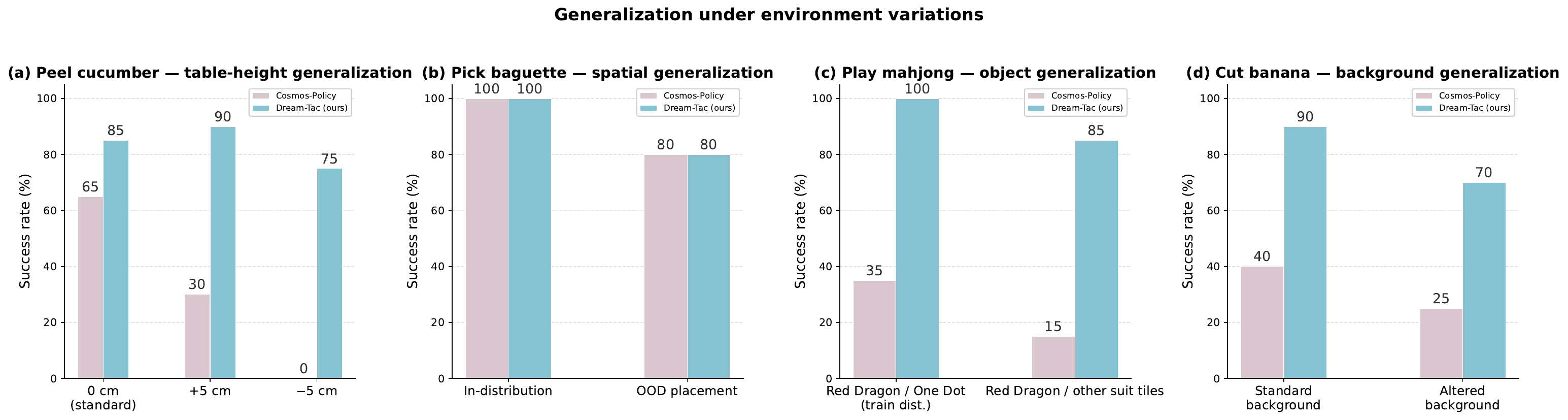}
    \caption{Generalization under environment variations. We compare Dream-Tac with Cosmos-Policy under four out-of-distribution settings: (a) table-height variation on \textit{Peel Cucumber}, (b) spatial variation on \textit{Pick Baguette}, (c) object variation on \textit{Play Mahjong}, and (d) background variation on \textit{Cut Banana}. Higher is better, and the numbers above the bars indicate success rates (\%).}
    \Description{Four bar charts comparing Dream-Tac and Cosmos-Policy under environment variations. Dream-Tac is more robust to table-height changes, tile appearance changes, and background changes, while both methods perform similarly under spatial placement variation.}
    \label{fig:generalization}
\end{figure*}

As shown in Fig.~\ref{fig:generalization}, Dream-Tac consistently outperforms Cosmos-Policy on three of the four settings and matches it on spatial generalization.
For \textbf{table-height generalization}, Dream-Tac achieves 85\%, 90\%, and 75\% success under the standard, +5\,cm, and -5\,cm settings, while Cosmos-Policy drops from 65\% to 30\% and 0\%.
For \textbf{object generalization} in \textit{Play Mahjong}, Dream-Tac improves success from 35\% to 100\% on training-distribution tiles and from 15\% to 85\% on unseen tile appearances.
For \textbf{background generalization} in \textit{Cut Banana}, Dream-Tac reaches 90\% and 70\% under the standard and altered backgrounds, substantially outperforming Cosmos-Policy (40\% and 25\%).
For \textbf{spatial generalization} in \textit{Pick Baguette}, both methods achieve the same performance, with 100\% success in-distribution and 80\% under unseen placements.

Overall, these results suggest that Dream-Tac generalizes better under variations that affect tactile-relevant interaction cues, while maintaining comparable performance under variations that are less related to tactile feedback.

\subsection{Training and Inference Efficiency}
\begin{figure}[h]
    \centering
    \includegraphics[width=0.5\textwidth]{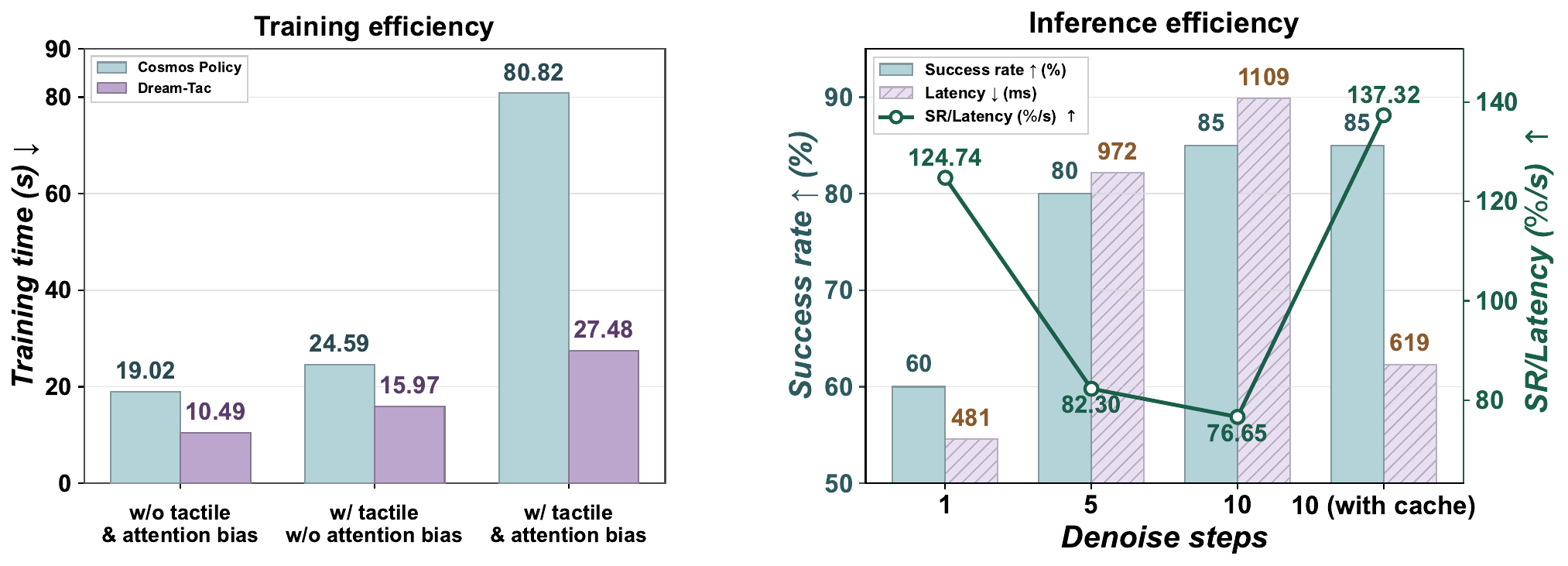}
    \caption{Training and inference efficiency of Dream-Tac on the \textit{Peel Cucumber} task. Left: training time comparison between Cosmos-Policy and Dream-Tac under three settings: without tactile input and attention bias, with tactile input but without attention bias, and with both tactile input and attention bias. Right: inference efficiency under different denoising steps and timestep caching, reporting success rate, latency, and success-rate-to-latency ratio (SR/Latency).}
    \label{fig:ab_step}
\end{figure}

\textbf{Training efficiency}
As shown in the left panel of Fig.~\ref{fig:ab_step}, Dream-Tac consistently improves training efficiency over Cosmos-Policy across all three settings.
Without tactile input and attention bias, Dream-Tac reduces training time from 19.02\,s to 10.49\,s, yielding a 44.8\% reduction (1.8$\times$ speedup).
When tactile tokens are introduced without attention bias, the training time is reduced from 24.59\,s to 15.97\,s, corresponding to a 35.1\% reduction (1.5$\times$ speedup).
The largest gain appears in the full setting with both tactile input and attention bias, where training time drops from 80.82\,s to 27.48\,s, corresponding to a 66.0\% reduction (2.9$\times$ speedup).

Notably, introducing tactile tokens alone leads to only moderate additional overhead for both methods, whereas enabling the structured attention bias causes a dramatic increase in training cost, especially for Cosmos-Policy.
Specifically, after adding attention bias on top of tactile modeling, the training time of Cosmos-Policy increases sharply from 24.59\,s to 80.82\,s, while Dream-Tac increases much more mildly from 15.97\,s to 27.48\,s.
This indicates that the dominant computational bottleneck does not come from tactile tokens themselves, but from the implementation of the structured logit bias.

The main reason is that our structured logit bias cannot be directly fused into standard FlashAttention \cite{dao2022flashattention}.
A naive implementation breaks the highly optimized fused attention path, which may require handling enlarged intermediate attention logits and incurs substantially higher memory traffic.
To address this, we reformulate the structured bias into a low-rank form and integrate it into blockwise attention computation, thereby avoiding materialization of the full attention matrix and reducing HBM accesses.
As a result, our efficiency optimization becomes especially effective in the full setting with both tactile input and attention bias, where the computational overhead of structured attention is the most significant.

\textbf{Effect of timestep cache}
As Fig.\ref{fig:ab_step} illustrates, incorporating denoising timestep caching substantially improves the inference efficiency of Dream-Tac for real-time control. Compared with full 10-step denoising, our accelerated system maintains the same 85\% success rate while reducing inference latency from 1109,ms to 619,ms, corresponding to a 
1.8$\times$ speedup. These results suggest that Dream-Tac can effectively benefit from timestep-level diffusion acceleration without sacrificing manipulation performance, making it significantly more practical for real-time robotic execution. Additional implementation details are provided in the appendix. 

\subsection{Analysis}



\subsubsection{Latent Separability of Tactile Representations}

\begin{figure}[h]
    \centering
    \includegraphics[width=0.42\textwidth]{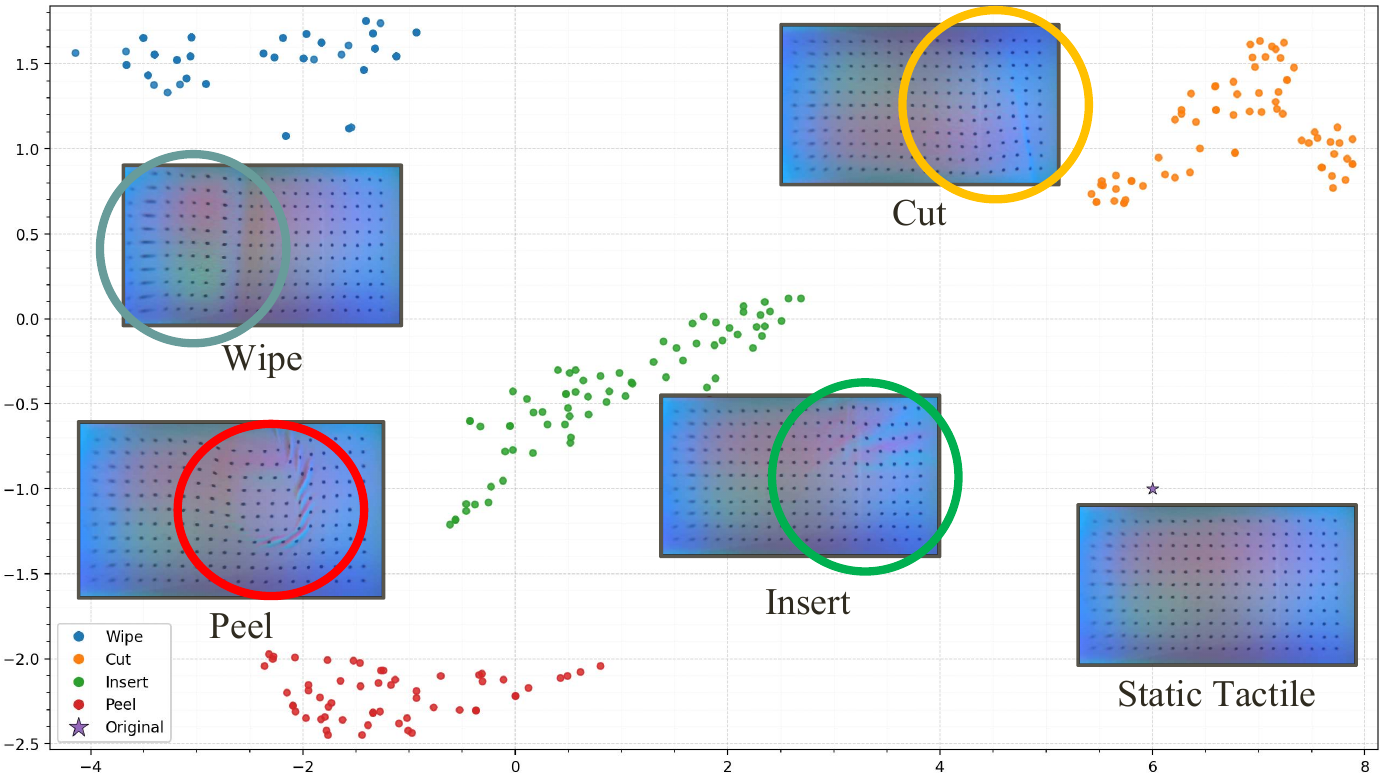}
    \caption{t-SNE visualization of tactile representations encoded by the pretrained Wan VAE. Different manipulation actions form distinct clusters in the latent space, and a representative tactile image is shown next to each cluster for reference.}
    \label{fig:t-sne}
\end{figure}
As Fig.~\ref{fig:t-sne} shows, the pretrained Wan VAE encoder is already sufficient to separate different tactile patterns in the shared latent space, despite their limited visual differences in pixel space. This result suggests that projecting tactile observations together with RGB images into a unified latent space does not destroy tactile-specific information, but instead preserves discriminative cues that are useful for manipulation. 

\subsubsection{Contact gate behavior}
\label{sec:analysis_gate}

The gate $g_t$ scales the tactile logit bias in Eq.~\eqref{eq:casa_logit}.
Here we analyze the gate \emph{offline} on randomly sampled training demonstrations from \textit{Peel Cucumber}.
For each timestep $t\!\ge\!1$, we compute the mean absolute RGB difference between frames $t{-}1$ and $t$ for each fingertip view, averaged over all pixels and channels and normalized by 255, and define $\rho_t$ as the maximum of the left and right values; we set $\rho_0\!=\!0$.
We then map $\rho_t$ to $g_t$ using Eq.~\eqref{eq:casa_gate} and the implementation hyperparameters stated there.

For visualization and summary statistics, we randomly sample five training episodes from the full training set, yielding 874 timesteps with $t\!\ge\!1$.
On this subset, $\rho_t$ has median $1.73\times10^{-3}$, 90th percentile $6.08\times10^{-3}$, mean $2.73\times10^{-3}$, standard deviation $2.32\times10^{-3}$, and coefficient of variation $\mathrm{std}/\mathrm{mean}\approx0.85$.
Within each episode, $g_t$ spans about $0.85$ of the $[g_{\min},g_{\max}]$ interval, while the time average $\frac{1}{T}\sum_t g_t$ ranges from $0.48$ to $0.61$ across the five episodes.
This indicates that many timesteps remain in a low-to-mid gate regime, while salient transients still drive $g_t$ over a large dynamic range.

\begin{figure}[t]
  \centering
  \includegraphics[width=\linewidth]{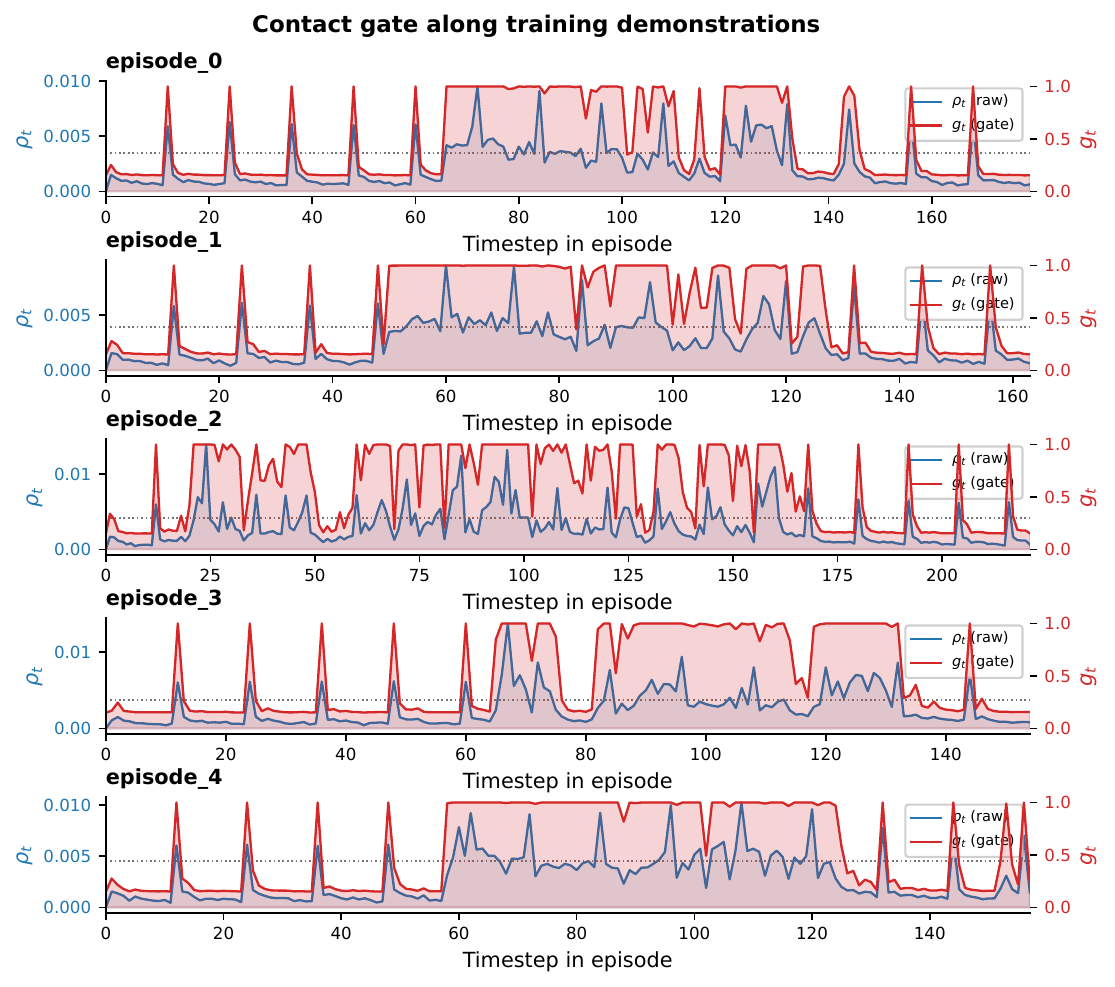}
  \caption{Tactile change $\rho_t$ (blue, left axis) and gate $g_t$ (red, right axis) over time, one panel per randomly sampled training episode from \textit{Peel Cucumber}. Dotted line: episode 75th percentile of $\rho_t$.}
  \Description{Stacked plots of rho and gate versus timestep for five randomly sampled training episodes. Peaks in rho align with increases in the gate, while flatter segments keep the gate lower.}
  \label{fig:gate_trajectories}
\end{figure}


Fig.~\ref{fig:gate_trajectories} shows that peaks in $\rho_t$---corresponding to rapid appearance changes in either tactile view---coincide with increases in $g_t$, while flatter segments keep the gate lower.
A closer episode-level inspection reveals a consistent temporal structure.
For \texttt{episode\_0}, \texttt{episode\_1}, \texttt{episode\_3}, and \texttt{episode\_4}, the early part of the trajectory exhibits a roughly periodic fluctuation in $g_t$.
This oscillatory pattern is likely caused by low-level tactile sensing noise and minor non-contact disturbances during the approach phase, when the gripper is moving toward the cucumber but stable contact has not yet been established.
Accordingly, during this pre-contact stage, the gate remains in a relatively low range, indicating that tactile cues are not yet strongly informative and should only weakly bias cross-modal attention.
Once contact occurs, $g_t$ rises rapidly and stays in a sustained high-value regime, suggesting that tactile information becomes substantially more relevant and should receive stronger emphasis during contact-rich interaction.
After the manipulation phase is completed and the contact intensity decreases, the gate returns to a lower level, consistent with the reduced importance of tactile feedback.

\texttt{episode\_2} follows a different pattern, with $g_t$ reaching a high regime already at the beginning of the trajectory.
This behavior is consistent with our data collection procedure, in which the robot initial pose is not strictly fixed across demonstrations.
For this episode, the recorded trajectory starts from a state already close to object contact, so tactile variation is elevated from the beginning and the gate is activated earlier.
Overall, these trajectories suggest that $g_t$ broadly tracks the manipulation phase: low during approach, high during sustained contact, and low again after the interaction subsides.

The fine-grained allocation of attention across tactile tokens remains governed by the content term $\mathbf{q}^\top\mathbf{k}$; $g_t$ only modulates, at a coarse level, how strongly non-tactile queries are biased toward tactile keys.

\section{Conclusion}
In this paper, We introduce Dream-Tac, a unified tactile world action model designed for contact-rich manipulation. Unlike prior world action models that rely mainly on visual observations, Dream-Tac jointly models actions, future visual observations, and tactile dynamics, enabling the model to better capture the physical interactions that are critical for manipulation. Our contact-gated visuo-tactile fusion and contact-aware attention bias allow the model to selectively emphasize tactile information during important contact transitions, while our dual-level acceleration design makes the framework more practical for real-time deployment. 
We evaluate Dream-Tac on six real-world contact-rich manipulation tasks, where it consistently outperforms strong baselines. In particular, Dream-Tac achieves higher success rates and generates more accurate future visual predictions, demonstrating the benefits of jointly modeling visuo-tactile dynamics. 
Overall, our results demonstrate the effectiveness of unified visuo-tactile world modeling and point to a promising direction for building more capable robotic systems in contact-rich environments.

\begin{acks}
This work was supported by the Beijing Natural Science Foundation (L252060).
\end{acks}

\newpage




%

\bibliographystyle{ACM-Reference-Format}
\bibliography{main}










\appendix
\clearpage

\section{Additional Experimental Details}
\subsection{Real-World Experimental Setup}
\label{app:real_world_setup}

\begin{center}
    \includegraphics[width=0.95\linewidth]{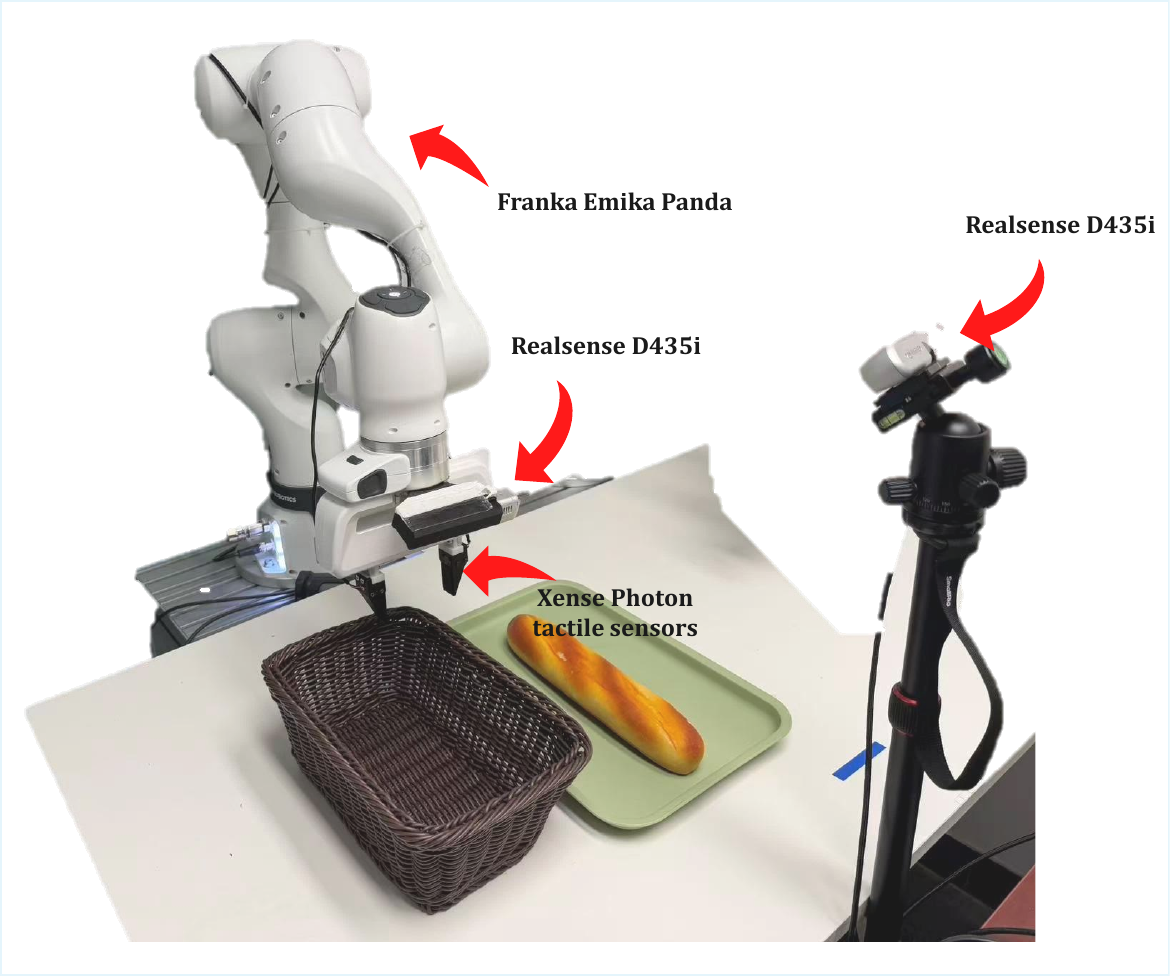}
    \captionof{figure}{\textbf{Real-world experimental setup.}
    Our platform consists of a Franka Emika Panda robot, a fixed third-person RealSense D435i camera, a wrist-mounted RealSense D435i camera, and two Xense Photon tactile sensors mounted on the gripper fingertips.}
    \label{fig:realworld_setup}
\end{center}

Fig.~\ref{fig:realworld_setup} illustrates the real-world platform used in our experiments.
The system consists of a Franka Emika Panda robot, a fixed third-person Intel RealSense D435i camera, a wrist-mounted Intel RealSense D435i camera, and two Xense Photon tactile sensors attached to the gripper fingertips.
This configuration provides complementary observations for contact-rich manipulation: the fixed camera captures the global scene layout, the wrist camera provides local egocentric views near the interaction region, and the fingertip tactile sensors capture contact and deformation patterns during manipulation.

\subsection{Task Details}
\noindent\textbf{Peel Cucumber}
The robot uses a peeler to remove the cucumber skin. Starting from one end of the cucumber, it establishes contact between the blade and the surface, and then slowly moves along the longitudinal direction toward the other end. During this motion, the robot maintains continuous contact with the curved surface while applying a consistent, moderate force to peel off a thin strip of skin. Tactile feedback plays a key role in this process, providing real-time signals about contact stability and resistance, which allows the robot to adjust its force and motion to avoid losing contact or cutting too deeply. Success is achieved if the cucumber skin is cleanly peeled off from the surface.

\noindent\textbf{Cut Banana}
The robot uses a knife to cut a banana. It first positions the blade above the target location and establishes contact with the banana surface, then applies a downward motion while maintaining a stable cutting direction. During the cutting process, the robot regulates the applied force to ensure smooth penetration without slipping or deviating from the intended trajectory. Tactile feedback provides signals about contact onset and resistance changes, allowing the robot to adjust force and maintain stable contact throughout the cut. Success is achieved if the knife cuts into the banana and creates a clear separation in the fruit.

\noindent\textbf{Insert USB}
The robot aligns a USB device with the port on a power strip and slowly inserts it along the correct direction. During insertion, it maintains stable contact and adjusts its motion to ensure proper alignment and smooth engagement. Tactile feedback provides critical signals about contact onset, alignment errors, and insertion resistance, enabling the robot to correct small misalignments and avoid excessive force. The task is considered successful if the USB is fully inserted into the port.

\noindent \textbf{Play Mahjong}
The robot observes the current tile and determines whether it completes a winning hand. Based on this decision, it executes the corresponding action: if the hand is winning, the robot pushes the tiles down to indicate a win; otherwise, it places the tile at the center of the table. During execution, the robot maintains stable contact with the tiles and the table surface, ensuring controlled pushing or precise placement. In this task, only tactile feedback is used by the robot to infer the current hand configuration, enabling it to determine whether the observed tile completes a winning hand. Success is achieved if the robot makes the correct judgment and executes the corresponding action accurately.

\noindent \textbf{Pick and Place}
The robot grasps a soft baguette from a plate and places it into a basket. It first approaches the baguette and establishes a stable grasp, then lifts it and transports it toward the basket before releasing it inside. Due to the deformable nature of the baguette, the robot needs to carefully regulate its grasp to avoid excessive deformation while maintaining sufficient support during lifting and transport. Tactile feedback reflects both object deformation and interaction dynamics, enabling the robot to adjust its grasp and motion to maintain stable control throughout the process. Success is achieved if the baguette is placed fully inside the basket without noticeably moving the basket.

\noindent \textbf{Clean Whiteboard}
The robot first grasps the eraser and brings it into contact with the board surface, then moves it across the marked area with a controlled wiping motion to remove the writing. During this process, the robot maintains continuous contact with the board while regulating the applied force to ensure effective cleaning without losing contact. After cleaning, it lifts the eraser and places it back at the designated location. Tactile feedback provides signals about contact consistency and friction during wiping, allowing the robot to maintain stable contact and effective interaction with the surface. Success is achieved if the writing is no longer visible and the eraser is correctly returned to its original position.

\begin{figure*}[h]
    \centering
    \includegraphics[width=\textwidth]{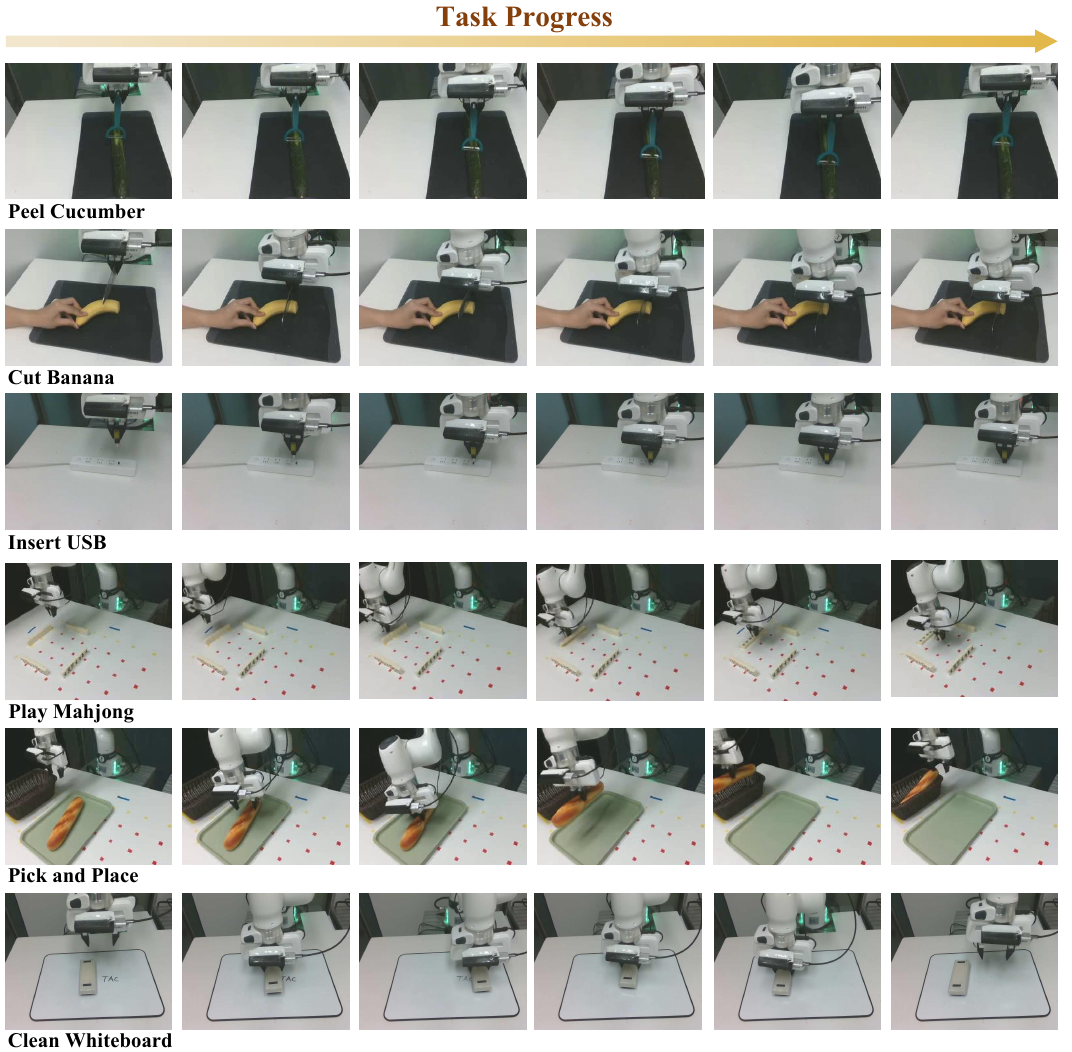}
    \caption{Visualization of Task Progress.}
    \label{fig:task_progress}
\end{figure*}

\subsection{Experimental Settings}
We use a 3D SpaceMouse to collect high-quality robot demonstrations. For each task, we collect 100 demonstration trajectories, while introducing a certain level of randomness in initial states and interactions to ensure diversity. We simultaneously record first-person and third-person camera observations, tactile images, and the robot’s proprioceptive states at a frequency of 30 Hz.

We list the parameters of expert demonstrations for different tasks in Table \ref{tab:parameters of task}. The length of each demonstration is not fixed and depends on the time required to complete the task.
“Demo” refers to the number of demonstrations collected for each task, “Episode Length” denotes the duration of each episode, “Teleop. Time” indicates the teleoperation time required to collect a single demonstration, and “Max Steps” represents the maximum execution steps allowed during evaluation.

During evaluation, each task is executed for 20 trials. The object positions are randomly initialized within a predefined range. A trial is considered a failure if the task is not successfully completed within the maximum allowed steps.    

\begin{table}[t]
    \renewcommand{\arraystretch}{1.2} 
    \centering
    \caption{\textbf{Parameters of expert demonstrations for real-world tasks.} ``Demo" refers to the number of demonstrations, ``Episode Length" denotes the duration of each episode in a task, ``Teleop. Times"  indicates the teleoperation time per demonstration, and ``Max Steps" represents the maximum execution time for a task during evaluation.}
    \label{tab:parameters of task}
    \footnotesize
    \resizebox{\linewidth}{!}{
        \begin{tabular}{c|cccc}
            \toprule
            \bf{Task Name} & \bf{Demos} & \bf{Avg. Episode Length} & \bf{Teleop. Times (s)} & \bf{Max steps} \\
            \midrule
            {\textbf{\textit{Peel Cucumber}}}
            &  100 & 220 & 7.3 & 450 \\
            {\textbf{\textit{Cut Banana}}}
            &  100 & 277 & 9.2 & 500 \\
            {\textbf{\textit{Insert USB}}}
            &  100 & 618 & 20.6 & 1100 \\
            {\textbf{\textit{Play Mahjong}}}
            &  100 & 200 & 6.7 & 400 \\
            {\textbf{\textit{Pick and Place}}}
            &  100 & 733 & 24.4 & 1200 \\
            {\textbf{\textit{Clean Whiteboard}}}
            &  100 & 833 & 27.8 & 1400 \\
            \bottomrule
        \end{tabular}
    }
    \vspace{-1em}
\end{table}

\subsection{Baseline Settings}
 We compare Dream-Tac with four strong baselines: (1) $\pi_0$ : a VLA flow model for general robot
control; (2) $\pi_{0.5}$, an enhanced VLA flow model for general physical reasoning and long-horizon manipulation; (3) ForceVLA : a multimodal VLA model integrating force-torque sensing for contact-rich manipulation; (4) Cosmos Policy : a world action model leveraging physical commonsense for complex task execution.

For $\pi_{0}$ and $\pi_{0.5}$ post-training, we use 1 NVIDIA H100 GPU across all tasks. We use both first-person and third-person camera observations as visual inputs. The action chunk size is set to 50. We train the model for 30,000 steps with a batch size of 32 and a learning rate of $5 \times 10^{-5}$.

For ForceVLA training, we use 8 NVIDIA H100 GPUs across all tasks. We train a separate model for each task using Adam with $(\beta_1,\beta_2)=(0.9,0.95)$ and a peak learning rate of $2.5 \times 10^{-5}$, decayed to $2.5 \times 10^{-6}$ over 30{,}000 steps. All runs use bfloat16 precision with gradient clipping at $\|\nabla\| = 1.0$.

For Cosmos-Policy training, we use the same training configuration as Dream-Tac across all tasks for a fair comparison. In particular, all Cosmos-Policy models are trained on 8 NVIDIA H100 GPUs with the same optimizer setting, learning-rate schedule, mixed-precision training strategy, and checkpoint-selection protocol as Dream-Tac.

\subsection{Training Hyperparameters}

We fine-tune Dream-Tac from the Cosmos-Predict2-2B Video2World checkpoint using Fused Adam with learning rate $10^{-4}$, $(\beta_1,\beta_2)=(0.9,0.99)$, $\epsilon=10^{-8}$, and weight decay $0.1$.
Training uses mixed bfloat16 precision.
The learning-rate multiplier is linearly warmed up over the first $2{,}000$ steps, decayed from $1.0$ to $0.3$ over the remainder of the first $20{,}000$ steps, and fixed at $0.06$ thereafter.

We adopt the rectified-flow / hybrid-EDM denoising objective inherited from the parent checkpoint, with $\sigma$ sampled from the default hybrid mixture: a log-normal component with $(p_\mu,p_\sigma)=(\ln 4,1.2)$ on $[0.01,200]$ and a uniform component on $[1,85]$.
Observed context latents remain clean ($\sigma=0$), while action and future-prediction tokens are jointly noised and denoised in a single forward pass.
Text dropout is set to $0$, and EMA is disabled.

We train on synchronized third-person and wrist RGB streams at $224\times224$, with proprioception and actions normalized to zero mean and unit variance.
For visuo-tactile models, synchronized left- and right-fingertip tactile RGB images are encoded with the same Wan VAE as the visual inputs and appended to the conditioning prefix.
Future tactile latents are jointly denoised with future visual latents and the action chunk, so tactile sensing serves as both an input modality and a prediction target.
Each sample uses a fixed chunk length $H=20$.
The per-GPU batch size is $25$ for the vision-only layout and $16$ for the visuo-tactile layout.

For models with contact-aware self-attention, we set the tactile logit-bias strength to $\alpha=2.0$ in Eq.~\eqref{eq:casa_logit}.
The gate hyperparameters $(m,s,k,\epsilon)$ and clipping rule follow Sec.~\ref{sec:analysis_gate} and Eq.~\eqref{eq:casa_gate}.
Checkpoints are saved periodically, and the final model is selected using the same validation-based rule across all methods.

\subsection{Contact Gate Statistics}
\label{subsec:gate_viz}

To better understand the behavior of the proposed contact gate, we visualize both the distribution of the tactile event strength $\rho_t$ and the deterministic mapping from $\rho_t$ to the gate value $g_t$.
As defined in Eqs.~\eqref{eq:casa_delta}--\eqref{eq:casa_gate}, the gate is computed directly from observed tactile RGB without introducing any learned gating network.
For each timestep $t \ge 1$, we compute the normalized mean absolute RGB difference between consecutive frames for the left and right fingertip views, and define $\rho_t$ as the maximum of the two values; for the initial timestep, we set $\rho_0=0$.
The resulting $\rho_t$ is then mapped to $g_t$ through the fixed sigmoid function in Eq.~\eqref{eq:casa_gate}, with $(m,s,k,\epsilon)=(0.002,0.001,4,10^{-6})$ and output range $[g_{\min}, g_{\max}] = [0.15, 1.0]$.

\begin{figure}[t]
  \centering
  \begin{subfigure}[t]{0.48\linewidth}
    \centering
    \includegraphics[width=\linewidth]{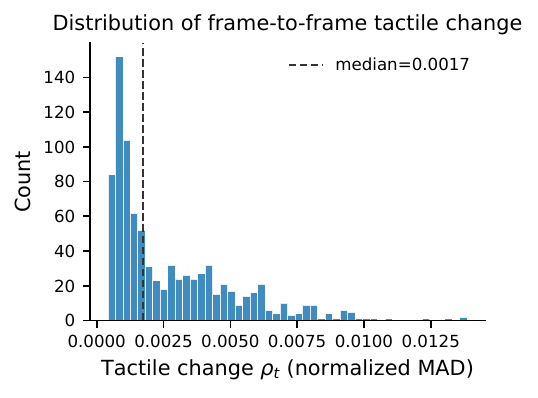}
    \caption{Distribution of frame-to-frame tactile change $\rho_t$.}
    \label{fig:gate_viz_hist}
  \end{subfigure}\hfill
  \begin{subfigure}[t]{0.48\linewidth}
    \centering
    \includegraphics[width=\linewidth]{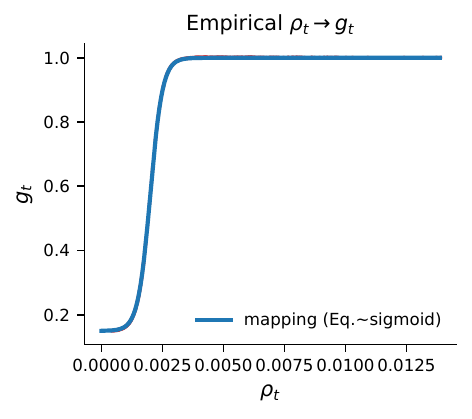}
    \caption{Deterministic mapping from $\rho_t$ to $g_t$.}
    \label{fig:gate_viz_curve}
  \end{subfigure}
  \caption{\textbf{Contact gate statistics.}
  (a) Empirical distribution of the tactile event strength $\rho_t$.
  (b) Mapping from $\rho_t$ to the gate value $g_t$ defined by Eq.~\eqref{eq:casa_gate}.}
  \Description{Two panels showing the distribution of tactile event strength and the corresponding mapping to the contact gate.}
  \label{fig:gate_viz}
\end{figure}

Fig.~\ref{fig:gate_viz} summarizes the resulting gate statistics.
Panel (a) shows the empirical distribution of $\rho_t$.
Most frame-to-frame tactile changes are small, corresponding to background fluctuations or non-contact phases, while larger values occur less frequently and are associated with salient interaction events.
Panel (b) shows the corresponding mapping from $\rho_t$ to $g_t$.
The gate remains low for small tactile changes and increases rapidly once $\rho_t$ exceeds the typical noise range, allowing the model to suppress weak tactile perturbations while amplifying tactile influence during contact transitions.
These results support the design of CASA: the gate is simple and deterministic, yet effectively separates quiescent periods from contact-relevant timesteps.

\section{Implementation Details of the Acceleration Modules}
\subsection{FlashBias Implementation}
\label{subsec:flashbias}

To efficiently implement the contact-aware attention bias in Eq.~\eqref{eq:casa_logit}, we adopt a FlashBias-style reformulation~\cite{wu2025flashbias} on top of fused scaled dot-product attention.
Recall that the tactile bias takes the form
\begin{equation}
\Delta_{ij}=\gamma_b\, a_i\, b_j,
\label{eq:rank_one_bias}
\end{equation}
where $\gamma_b$ absorbs the gate value and fixed scale, and $a_i,b_j\in\{0,1\}$ indicate the query/key positions involved in the asymmetric tactile bias.
This bias is rank-one and can therefore be written as $\Delta_{ij}=u_i v_j$, with $u_i=\sqrt{\gamma_b}\,a_i$ and $v_j=\sqrt{\gamma_b}\,b_j$.

Instead of explicitly materializing a dense $S\times S$ bias matrix, we fold this term into standard dot-product attention by augmenting queries and keys with one additional scalar channel:
\begin{equation}
\widetilde{\mathbf{q}}_i=
\Big[\frac{1}{\sqrt{d}}\mathbf{q}_i \;\Vert\; u_i\Big],
\qquad
\widetilde{\mathbf{k}}_j=
\Big[\mathbf{k}_j \;\Vert\; v_j\Big].
\label{eq:qk_tilde}
\end{equation}
Their inner product then becomes
\begin{equation}
\langle \widetilde{\mathbf{q}}_i,\widetilde{\mathbf{k}}_j\rangle
=
\frac{\mathbf{q}_i^\top \mathbf{k}_j}{\sqrt{d}} + u_i v_j,
\label{eq:logit_equivalence}
\end{equation}
which is exactly equivalent to adding the structured bias $\Delta_{ij}$ to the original attention logits.

This reformulation allows us to compute attention with a fused scaled dot-product attention operator on $(\widetilde{\mathbf{Q}},\widetilde{\mathbf{K}},\mathbf{V})$ without introducing a dense additive mask.
When necessary, we pad the augmented query and key channels with zeros to satisfy kernel alignment requirements; this does not affect the resulting logits.
Compared with a naive dense-bias implementation, this design preserves the fused attention path and avoids explicit $O(S^2)$ bias materialization, while increasing the per-token state by only $O(1)$ extra channels per head.

In practice, this FlashBias-based implementation substantially improves training efficiency for Dream-Tac, since the contact-aware bias can be incorporated without breaking the optimized fused attention kernel.

\subsection{Diffusion-Step Time Cache}

\paragraph{Diffusion Step Cache.}

In this section, we briefly introduce diffusion step cache. Let \(x_t\) denote the noisy latent at diffusion step \(t\), and let the denoising model be \(\epsilon_\theta(x_t, t, c)\), where \(c\) is the condition. A standard reverse step is written as
$$
x_{t-1} = \Phi\!\left(x_t, \epsilon_\theta(x_t, t, c)\right),
$$
where \(\Phi(\cdot)\) is the scheduler update.

Diffusion step cache is based on the observation that adjacent denoising steps often produce similar intermediate features. Suppose the model can be decomposed as
\[
\epsilon_\theta(x_t, t, c) = h_\theta\!\left(g_\theta(x_t, t, c)\right),
\]
where \(g_\theta\) is an expensive intermediate computation. Instead of recomputing \(g_\theta\) at every step, we cache its output at step \(t_k\),
\[
z_{t_k} = g_\theta(x_{t_k}, t_k, c),
\]
and reuse it for nearby steps \(t \approx t_k\) by
\[
g_\theta(x_t, t, c) \approx z_{t_k}.
\]
Thus, the denoising prediction becomes
\[
\epsilon_\theta(x_t, t, c) \approx h_\theta(z_{t_k}),
\]
and the reverse process is approximated as
\[
x_{t-1} \approx \Phi\!\left(x_t, h_\theta(z_{t_k})\right).
\]

In this way, diffusion step cache reduces repeated computation across similar denoising steps, improving sampling efficiency with limited approximation error.

\begin{figure}[t]
    \centering
    \includegraphics[width=\columnwidth]{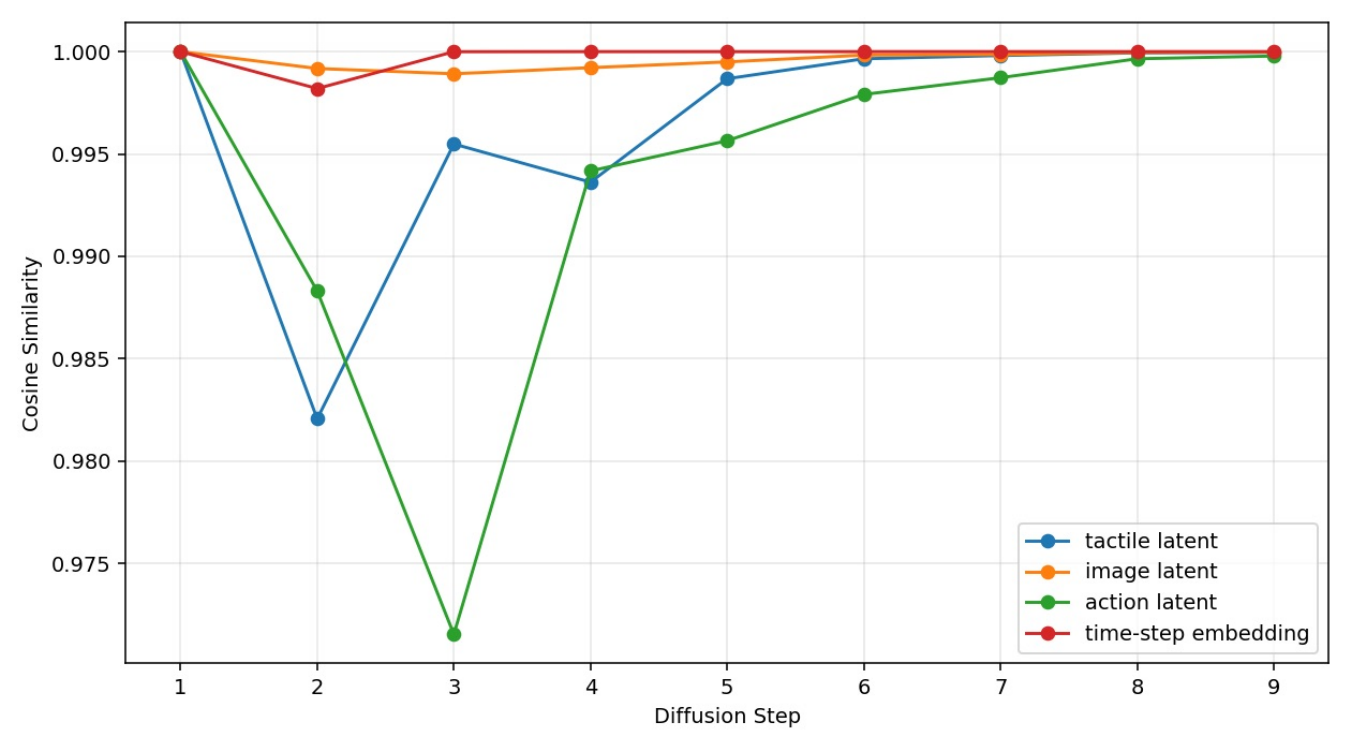}
    \vspace{0.5mm}
    
    \includegraphics[width=\columnwidth]{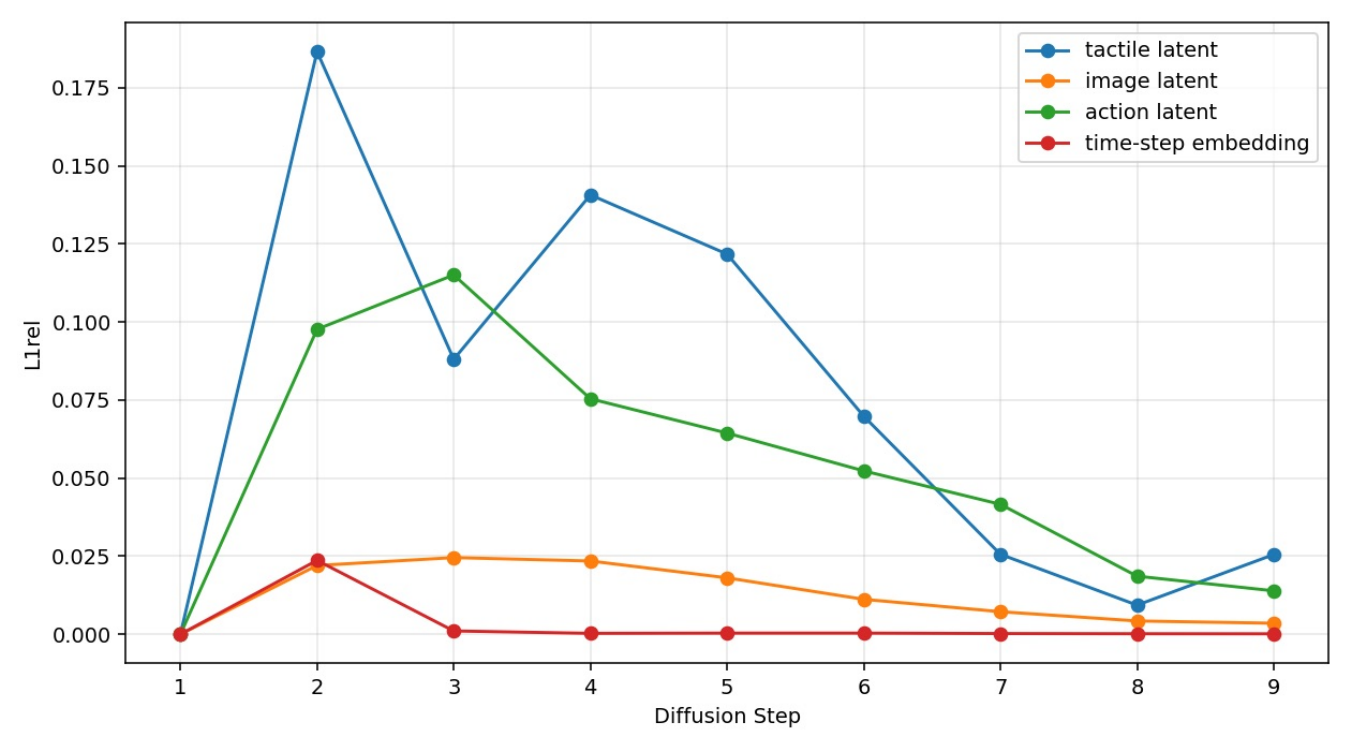}
    \vspace{-2mm}
    \caption{Comparison of two diffusion-step similarity metrics. Top: cosine similarity. Bottom: relative \(L_1\) distance.}
    \label{fig:time-cache}
\end{figure}

Some recent diffusion step cache methods further design indicators to estimate whether the model output changes significantly across steps. For example, TeaCache \cite{liu2024timestep} shows that the timestep-embedding-modulated noisy input can serve as an effective indicator of output variation, and uses it to decide whether cached results can be reused at the current step. In this way, diffusion step cache reduces repeated computation across similar denoising steps and improves sampling efficiency with limited approximation error. 
To facilitate this technique, we further analyze the redundancy across diffusion steps by visualizing the open-loop forward inference results on the validation set. As shown in the cosine-similarity and relative-\(L_1\) plots in Figure \ref{fig:time-cache}, the variation of action latents across diffusion steps is not consistently reflected by the timestep embedding. Specifically, neither relative \(L_1\) distance nor cosine similarity between the timestep embedding and latent features provides a reliable indicator of action change. This suggests that using timestep embedding as a proxy for output variation is not well justified in our setting. 

At the same time, we observe strong redundancy between adjacent diffusion steps. In particular, the cosine similarity between neighboring steps remains consistently high, with an average value of approximately \(0.997\), indicating that the model outputs change only marginally across most denoising steps. This high similarity motivates the use of a cache-based acceleration strategy during inference. 

Based on this observation, we perform full forward computation only at the first step and at the third step, which exhibits the largest variation on the validation set, and reuse cached results for all remaining steps. This design preserves generation quality while substantially reducing inference cost.

\section{Additional Reconstruction Results}

To further qualitatively evaluate the predictive capability of Dream-Tac, we provide additional reconstruction results for both visual reconstruction and joint visuo-tactile reconstruction.
These results complement the main paper by showing that Dream-Tac not only improves downstream manipulation performance, but also learns temporally coherent and physically meaningful future representations across modalities.

\subsection{Visual Reconstruction}

\begin{figure*}[t]
    \centering
    \includegraphics[width=0.65\textwidth]{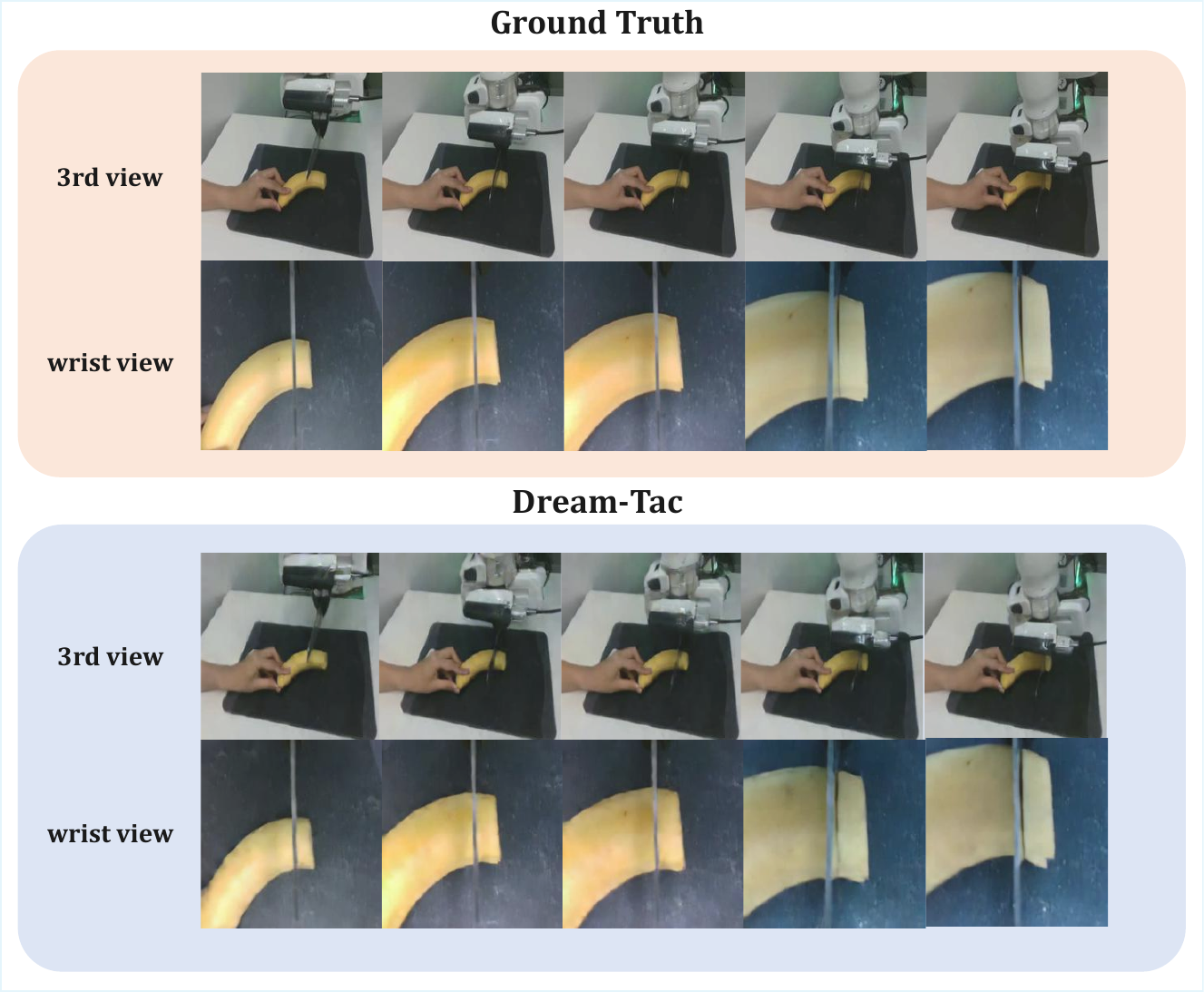}
    \caption{Comparison of Dream-Tac-predicted future images with ground-truth images on the \textit{Cut Banana} task.}
    \label{fig:vis}
\end{figure*}

Fig.~\ref{fig:vis} presents representative visual reconstruction results on the \textit{Cut Banana} task, comparing Dream-Tac-predicted future images with the corresponding ground-truth images.
Overall, Dream-Tac is able to faithfully capture the global scene layout, the relative geometry of the robot and manipulated object, and the coarse motion trend over future timesteps.
In particular, the predicted frames preserve the spatial relationships among the gripper, knife, and banana, indicating that the model has learned a stable action-conditioned representation of future visual dynamics.

Although some fine-grained textures and high-frequency details are inevitably smoothed, the predicted results remain semantically consistent with the ground truth and correctly reflect the key progression of the manipulation process.
This suggests that Dream-Tac does not merely generate visually plausible frames, but learns predictive visual structures that are sufficiently aligned with the underlying task evolution.
Such predictive ability is important for world action models, where future visual reconstruction serves as an informative intermediate representation for action generation.

\subsection{Visuo-Tactile Reconstruction}

Beyond visual reconstruction alone, Dream-Tac jointly models future visual observations and future tactile observations in a shared latent space.
This joint reconstruction behavior is particularly important in contact-rich manipulation, where critical state transitions are often only weakly observable in RGB images but are clearly reflected in tactile signals.
By reconstructing both modalities together, Dream-Tac is encouraged to learn future representations that are not only visually coherent but also physically grounded.

\begin{figure*}[t]
    \centering
    \includegraphics[width=0.62\textwidth]{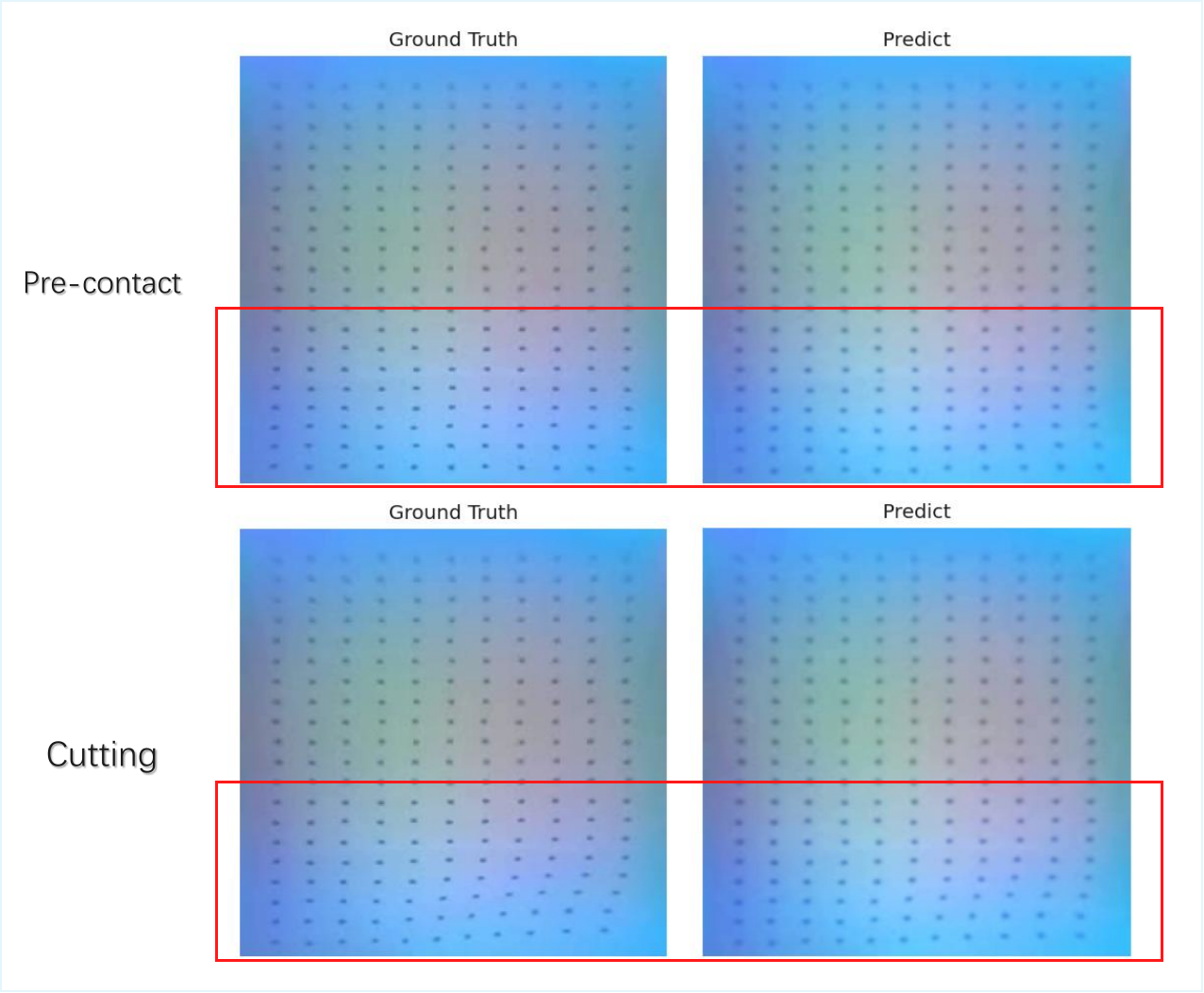}
    \caption{Comparison of Dream-Tac-predicted future tactile observations with ground-truth tactile observations on the \textit{Cut Banana} task.
    The top row shows a \textit{Pre-contact} phase, where the predicted tactile image remains close to the undeformed background pattern.
    The bottom row shows a \textit{Cutting} phase, where the predicted tactile image captures the contact-induced deformation pattern and intensity change observed in the ground truth.}
    \label{fig:tactile_vis}
\end{figure*}

Fig.~\ref{fig:tactile_vis} shows representative tactile reconstruction results from the \textit{Cut Banana} task under two phases: \textit{Pre-contact} and \textit{Cutting}.
In the \textit{Pre-contact} phase, the predicted tactile image remains highly consistent with the ground truth and preserves the nearly undeformed background pattern, indicating that Dream-Tac does not hallucinate spurious contact signals before physical interaction occurs.
In the \textit{Cutting} phase, the predicted tactile image reproduces the overall deformation pattern and intensity redistribution observed in the ground truth, especially in the lower contact-sensitive region where the tactile response becomes more pronounced.
Although some fine-grained pin-level details are slightly smoothed, the prediction still captures the main spatial structure of the contact pattern and its phase-dependent change from non-contact to sustained interaction.

These results suggest that Dream-Tac captures meaningful cross-modal dynamics rather than treating touch as an auxiliary side signal.
In particular, the model can distinguish quiescent tactile states from contact-active tactile states and predict how tactile feedback evolves together with the manipulation process.
This provides additional evidence for our central design motivation: in contact-rich manipulation, predicting how the future interaction will \emph{feel} is as important as predicting how the future scene will \emph{look}.
The tactile reconstruction results therefore help explain why Dream-Tac achieves stronger performance on tasks that require precise contact perception and fine-grained physical interaction.

\section{Limitations and Future Work}

Despite the promising results, our work has several limitations. 
First, Dream-Tac is evaluated on a limited set of real-world contact-rich manipulation tasks, and its generalization to broader task families, more diverse objects, and more complex environments remains to be further verified. 
Second, although the proposed contact-aware design improves tactile utilization, the current gating mechanism is still based on relatively simple frame-to-frame tactile variation and may not fully capture more subtle or long-horizon interaction patterns. 
Third, while our efficiency optimizations improve practical deployment, diffusion-based world action models remain computationally expensive compared with lightweight reactive policies.

In future work, we plan to scale Dream-Tac to a larger and more diverse visuo-tactile dataset, explore more expressive contact modeling mechanisms, and improve efficiency for faster real-time control. 
Another important direction is to extend the framework to more challenging settings, such as dexterous multi-stage manipulation, deformable object interaction, and long-horizon contact-rich tasks.

\end{document}